\documentclass[journal]{IEEEtran}
\usepackage{multirow}
\usepackage{cite}
\usepackage{amsmath,amssymb,amsfonts}
\usepackage{graphicx}
\usepackage{textcomp}
\usepackage{algorithm}  
\usepackage{algpseudocode}  
\usepackage{amsmath}  
\usepackage{booktabs}
\usepackage{array}
\usepackage{url}

\usepackage{longtable}
\usepackage{supertabular,booktabs}

\newcommand{\tabincell}[2]{\begin{tabular}{@{}#1@{}}#2\end{tabular}}
\graphicspath{{Figs/}}
\begin{document}

\title{Review of data analysis in vision inspection of power lines with an in-depth discussion of deep learning technology}

\author{
	Xinyu Liu,
    Xiren Miao,
    Hao Jiang, \IEEEmembership{Member, IEEE},
    Jing Chen
\thanks{
	This work was supported
	in part by the Key Natural Foundation for Young Scholars of Fujian Province under Grant JZ160415,
	in part by the Research Program of Distinguished Young Talents of Fujian Province under Grant 601934,
	in part by the National Natural Science Foundation of China under Grant 61703105 and Grant 61703106, 
	in part by the Natural Science Foundation of Fujian Province under Grant 2017J01500,
	in part by the Qishan Talent Support Program of Fuzhou University under Grant XRC-1623,
	and in part by the Research Foundation of Fuzhou University under Grant XRC-17011.
	(Corresponding author: Hao Jiang)
	}
\thanks{
	X. Liu, X. Miao, H. Jiang and J. Chen are with the College of Electrical Engineering and Automation, Fuzhou University, Fuzhou 350108, China e-mail: (xinyu3307@163.com; miaoxr@163.com; jiangh@fzu.edu.cn; chenj@fzu.edu.cn).
	}
}


\maketitle

\begin{abstract}
The widespread popularity of unmanned aerial vehicles enables an immense amount of power lines inspection data to be collected. 
How to employ massive inspection data especially the visible images to maintain the reliability, safety, and sustainability of power transmission is a pressing issue.
To date, substantial works have been conducted on the analysis of power lines inspection data. 
With the aim of providing a comprehensive overview for researchers who are interested in developing a deep-learning-based analysis system for power lines inspection data, this paper conducts a thorough review of the current literature and identifies the challenges for future research.
Following the typical procedure of inspection data analysis, we categorize current works in this area into component detection and fault diagnosis.
For each aspect, the techniques and methodologies adopted in the literature are summarized.
Some valuable information is also included such as data description and method performance.
Further, an in-depth discussion of existing deep-learning-related analysis methods in power lines inspection is proposed.
Finally, we conclude the paper with several research trends for the future of this area, such as data quality problems, small object detection, embedded application, and evaluation baseline.

\end{abstract}


\IEEEpeerreviewmaketitle

\section{Introduction}
\IEEEPARstart{P}{ower} lines inspection for uninterrupted supply has become an important topic due to the increasing dependency of modern-day societies on electricity.
The power line is established by several types of components with different function that include insulator, tower, conductor and fitting.
Due to out-door environment in complex landform and volatile weather, the power line component could be damaged frequently.
One faulty component (e.g., conductor fault), or generally the combination of multiple damaged components (e.g., fitting faults) can cause a power outage.
Once the power lines are malfunction in one region, it may lead to supra-regional blackouts and may even cause catastrophic accidents such as fire in forest area \cite{wang2019danerous}.
The objective of power lines inspection is to check the condition of the power line component，
And then, the inspection result as a guide is used for power companies to decide which component should be maintained or replaced.
A fast and accurate inspection can greatly increase the efficiency of maintenance decision-making, and further reduce the possibility of power line failures, which is the guarantee of safe and reliable power supply \cite{qin2017fault_causes}.

However, the power lines inspection is facing several challenging problems such as extensive area, a large number of components, and complex natural environments.
Traditional inspection methods including manual ground survey and helicopter-assisted patrol which have been used for decades \cite{liu2019ground_aerial_inspection}.
Both methods inspect the power lines by human visual observations, which are high cost, high risk, low efficiency, and long-term operating \cite{matikainen2016review_remote}.
In recent years, the development of Unmanned Aerial Vehicle (UAV) and digital image technologies provides a new platform for power lines inspection \cite{shakhatreh2019uav_review}.
The UAV inspection method separates the traditional inspection into two parts: data collection and data analysis.
The inspector remotely operates the UAV to collect images for inspection targets, and then the captured images or videos are sent to workers who have professional skills for data analysis.
Due to the advantages of low cost, high security and high efficiency, deploying UAV inspection to replace traditional methods which is based on manual labor has been tried extensively.

UAV inspection as a recent method greatly reduces the work intensity of inspectors and improves the efficiency of power lines inspection, but it also brings massive data.
In addition, these images and videos are usually analyzed by a time-consuming manual approach which is expensive, potentially dangerous and not enough accurate \cite{martinez2018big_data}.
In the past few years, many researchers have been seeking to develop fast and accurate analysis methods to automatically recognize the condition of power lines in aerial images \cite{nguyen2018review_dl_inspection}.
These researches cover a wide range of power line components and their faults with various image processing technologies.
Moreover, most of them are task-specific that focusing on one particular component or fault.
The main objective of this paper is to provide the state-of-the-art of vision-based inspection of power line components in research literature, and to present some degree of taxonomy that gives readers a helpful accessible understanding of similarities and differences between a wide variety of studies.
We aim to offer an overview of the possibilities and challenges provided by modern computer vision technology from the perspective of inspection data analysis to discuss the potential and limitations of different analysis methods.
Note that the visible images captured from UAVs are the most commonly used in power lines inspection due to their low cost, humanized observation and detailed information.
Therefore, in this review paper, we only consider the analysis method of visible images while works about other data sources and the procedure of data collection are not included.

In this paper, we first provide some related works of vision-based power lines inspection from the perspective of data analysis.
The bibliometrics analysis of the literature, relevant review articles, datasets for public and the taxonomy used in this paper are included.
Next, we introduce several basic concepts in power lines inspection.
These concepts contain inspection method and data source with spacial attention paid to UAV inspection and visible images, and main components with their roles and common faults.
Then, we review the studies found in literature of analysis methods of visible images in power lines inspection.
These research articles mainly published in the past five years, are summarized into two categories including component detection and fault diagnosis.
The main ideas of the analysis method, description of the dataset, and some representative quality analysis results are presented to understand the capabilities of various analytic approaches in different applications.
Based on that, we propose an in-depth discussion of deep-learning-related methods in the researches reviewed above.
A brief introduction of fundamental deep learning technologies, the exploration of analysis methods related to deep learning, and a basic conception of inspection data analysis system using several alternative image processing approaches are presented.
Finally, we discuss open research questions for future research directions.

The remainder of this paper is organized as follows.
Section II provides the related works.
Section III offers a brief introduction of power lines inspection including inspection method, data source, and power line components with their common faults.
Section IV conducts the survey on inspection data analysis from the perspective of component detection and fault diagnosis.
Section V presents the in-depth discussion of the analysis methods reviewed in Section IV that are deep-learning-related.
Section VI discusses the open research issues.
Section VII draws the conclusion.

\section{Related works}
\subsection{Bibliometric Analysis}
In order to provide an overview of the existing research in vision-based inspection of power lines, a bibliometric analysis was conducted on 9 December 2019 using the acknowledged databases, Google Scholar. 
The query for Google Scholar is as follows: power AND (visual OR image* OR vision) AND (aerial OR UAV* OR overhead) AND "power line *". 
Intend to further screen out researches related to deep learning, an extended query is applied: power AND (visual OR image* OR vision) AND (aerial OR UAV* OR overhead) AND "power line*" AND "deep learning".

\begin{figure}[ht]
	\centering
	\includegraphics[width=8.5cm]{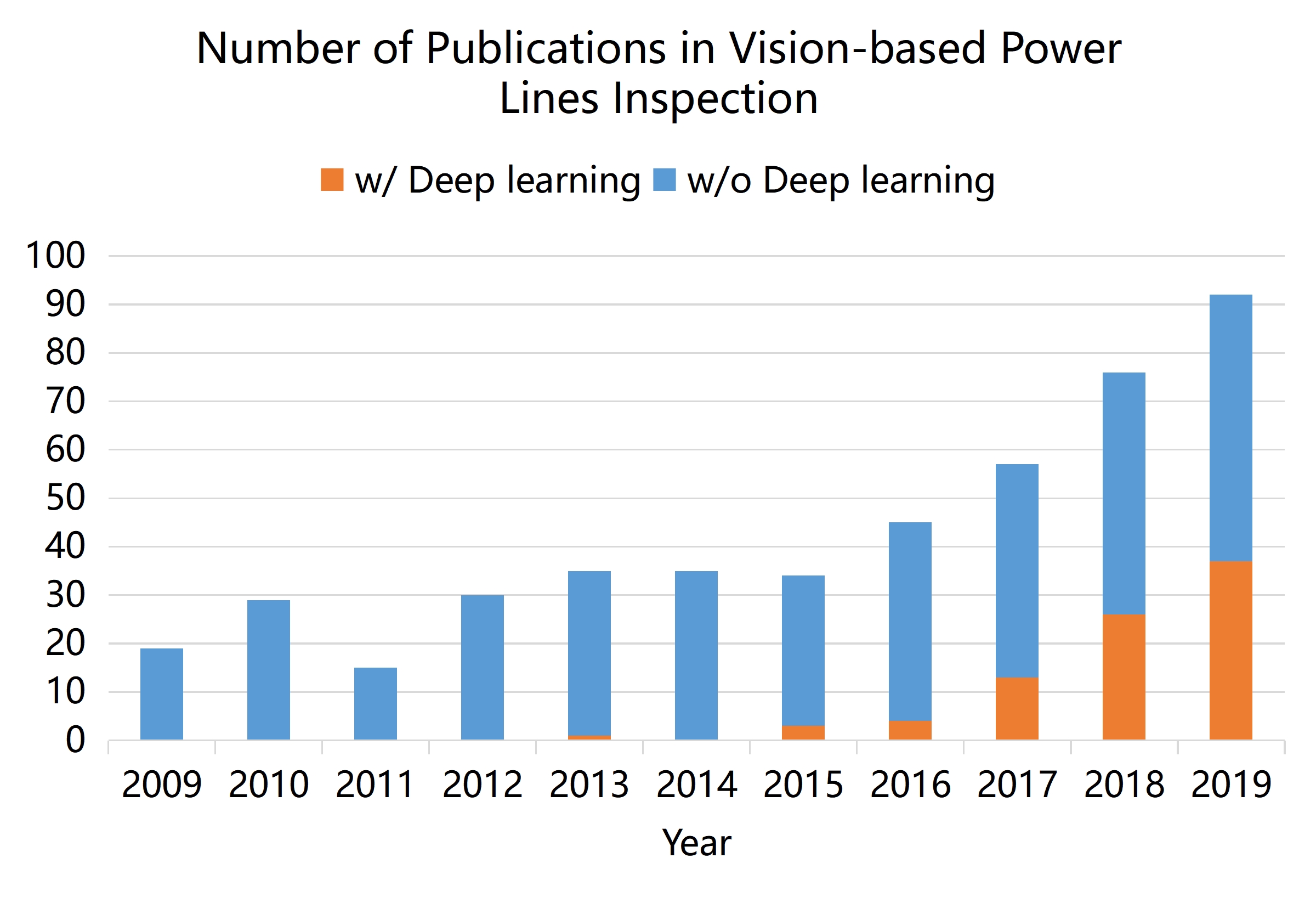}
	\caption{Number of publications indexed by Google Scholar.}\label{fig:num_pub_w_DL}
\end{figure}

\mbox{Fig. \ref{fig:num_pub_w_DL}} illustrates the number of publications indexed by Google Scholar from 2009 to 2019. 
Totally 477 research articles can be found that include 84 publications related to deep learning.
Before 2015, the number of total publications was at a relatively low and stable level.
Since 2016, researches in vision-based power lines inspection increased yearly and reached the number of 92 in the year of 2019.
As early as 2013, there was a research mentioned about "deep learning" but the deep learning technology didn't really apply until 2016.
These deep-learning-based publications also increased year by year since 2016 and reached 37 in the year of 2019.
This result should not be a surprise that aerial inspection just have been widely applied by power companies in recent years with the development of UAV and deep learning technologies. 
It takes times for power companies to collect inspection data and for researchers to design and evaluate their methods in a specific real-world application.

\subsection{Relevant Review Articles}
%
Several review articles related to power line inspection have been published in the past decade.
Some of them focused on inspection platforms.
Katrasnik et al. \cite{katrasnik2009survey_robots} presented the achievements in power line inspection by mobile robots including flying robots and climbing robots.
Toussaint et al. \cite{toussaint2009review_robots} conducted a review of power line inspection and maintenance which focused on climbing robots designed to cross obstacles.
Tong et al. \cite{Tong2010review_helicopter} summarized the image processing based applications in power line inspection by helicopter.
A few review articles discussed the specific application of power line inspection, which focused on one kind of the component or fault.
Ahmad et al. \cite{ahmad2013review_vegetation} proposed a review of advantages and limitations related to the vegetation encroachment monitoring of power lines
Prasad et al. \cite{prasad2016review_insulator} discussed the vision-based techniques for insulator monitoring of power lines.
With the development of sensor technique, a number of remotely sensed data sources were applied in power line inspection.
Mirall{\`e}s et al. \cite{miralles2014review_cv_inspection} conducted a review of several vision-based applications in the management of power lines with respect to different vision sensors.
Matikainen et al. \cite{matikainen2016review_remote} presented a remote sensing-based survey of power lines and their surroundings in research literature. 
A wide range of data sources was discussed from coarse satellite images to detailed visible images.

Deep learning has achieved great success in computer vision since 2012, but the deep learning based application in power lines inspection was not reported until 2016.
Nguyen et al. \cite{nguyen2018review_dl_inspection} conducted a literature review of automatic vision-based inspection of power lines which aimed to discuss the role and possibilities of deep learning technique. 
They summarized the existing researches of the vision-based power line inspection from the perspectives of UAV navigation and inspection task. 
However, the research reviewed in \cite{nguyen2018review_dl_inspection} is mostly pre-2018 when the deep learning was hardly applied to power lines inspection at that time. 
Hence, they proposed a potential concept of automatic power line inspection system based on deep learning rather than reviewing the research articles.

The review papers mentioned above summarized the researches of power lines inspection from different aspects including inspection platforms\cite{katrasnik2009survey_robots, toussaint2009review_robots,Tong2010review_helicopter}, specific inspection applications\cite{ahmad2013review_vegetation, prasad2016review_insulator}, inspection data sources \cite{miralles2014review_cv_inspection, matikainen2016review_remote} and automatic inspection systems\cite{nguyen2018review_dl_inspection}.
Our paper differs from the above reviews especially reference \cite{nguyen2018review_dl_inspection} by only focusing on component inspection task of power lines rather than including data collection.
Special attention is paid to visible image analysis based on deep learning.
In addition, an in-depth exploration of the analysis methods that are aiming at components detection and their faults diagnosis are provided.
Beside that, after years of development, more novel methods are proposed and more challenges are defined. 
The existing reviews prior to the recent striking success are not as up-to-date as this paper.
We give more emphasis on the research over the past five years while typical works that were published earlier are also included. 

\subsection{Datasets for public}
Due to the confidentiality of the inspection data of power lines, most of the power companies are hesitant to make their data public available.
This results in research challenges such as data insufficiency and missing evaluation baseline.
Nevertheless, there are several datasets offered by personal researchers that have been released to the public over the past few years.
Here, we summarize these datasets with brief description and provide their website.
\begin{itemize}
	\item 
	\textbf{Insulator dataset in reference \cite{tao2018ILN_DDN_ins}}: 
	The dataset consists of 848 aerial images that the main object in this dataset is the insulator in power lines.
	Totally 600 of them are captured in real-world and labeled with insulator.
	The rest images are synthetized by hand and labeled with insulator fault, in particular the missing-cap fault of insulator.
	\item 
	\textbf{Tower dataset in reference \cite{bian2019Fst_tower}}:
	There are 1300 images in this dataset, and the major object is electrical tower.
	Most of the images are collected from the inspection video and the internet.
	Various kinds of tower with different backgrounds are included. 
	\item 	
	\textbf{Conductor dataset in reference \cite{lee2017weakly_line}}:
	This dataset contains totally 8400 images collected from visible and infrared cameras in equal quantity.
	To achieve multi scale recognition, images with close and far scene are included.
	In addition, the dataset is separated into two sub-sets according to different annotations for weakly supervised learning.
	Sub-set 1 is labeled with image level annotations which has 8000 images while another is labeled with pixel level annotations.
\end{itemize}

\begin{table*}[ht]
	\centering
	\caption{Basic information of several open inspection datasets}
	\label{tab:open_dataset}
	\renewcommand\arraystretch{1.2}
	\begin{tabular}{ c l c m{7cm}}
		\hline
		\textbf{Dataset} & \textbf{~~~~~~~~~~~~~~~Brief Description} & \textbf{Quantity} & \textbf{~~~~~~~~~~~~~~~~~~~~~~~~~~~Website} \\
		\hline
		Insulator \cite{tao2018ILN_DDN_ins} & \tabincell{l}{Real-world images labeled with insulator \\ Synthetic images labeled with defect (missing-cap)} & 848 & \url{https://github.com/InsulatorData/InsulatorDataSet}  \\ 
		\hline
		Tower \cite{bian2019Fst_tower} & \tabincell{l}{Collected from internet and inspection videos \\ Various types of towers and backgrounds} & 1300 & \url{https://drive.google.com/drive/folders/1UyP0fBNUqFeoW5nmPVGzyFG5IQZcqlc5}  \\ 
		\hline
		Conductor \cite{lee2017weakly_line} & \tabincell{l}{Captured by visible and infrared cameras \\ Sub-set1 labeled with image level annotations \\
			Sub-set2 labeled with pixel level annotations} & 8400 & \tabincell{l}{Dataset1:\url{https://data.mendeley.com/datasets/n6wrv4ry6v/3} \\ Dataset2:\url{https://data.mendeley.com/datasets/twxp8xccsw/6}} \\ 
		\hline
	\end{tabular}
\end{table*}

\subsection{Taxonomy}
\begin{figure}[ht]
	\centering
	\includegraphics[width=8.5cm]{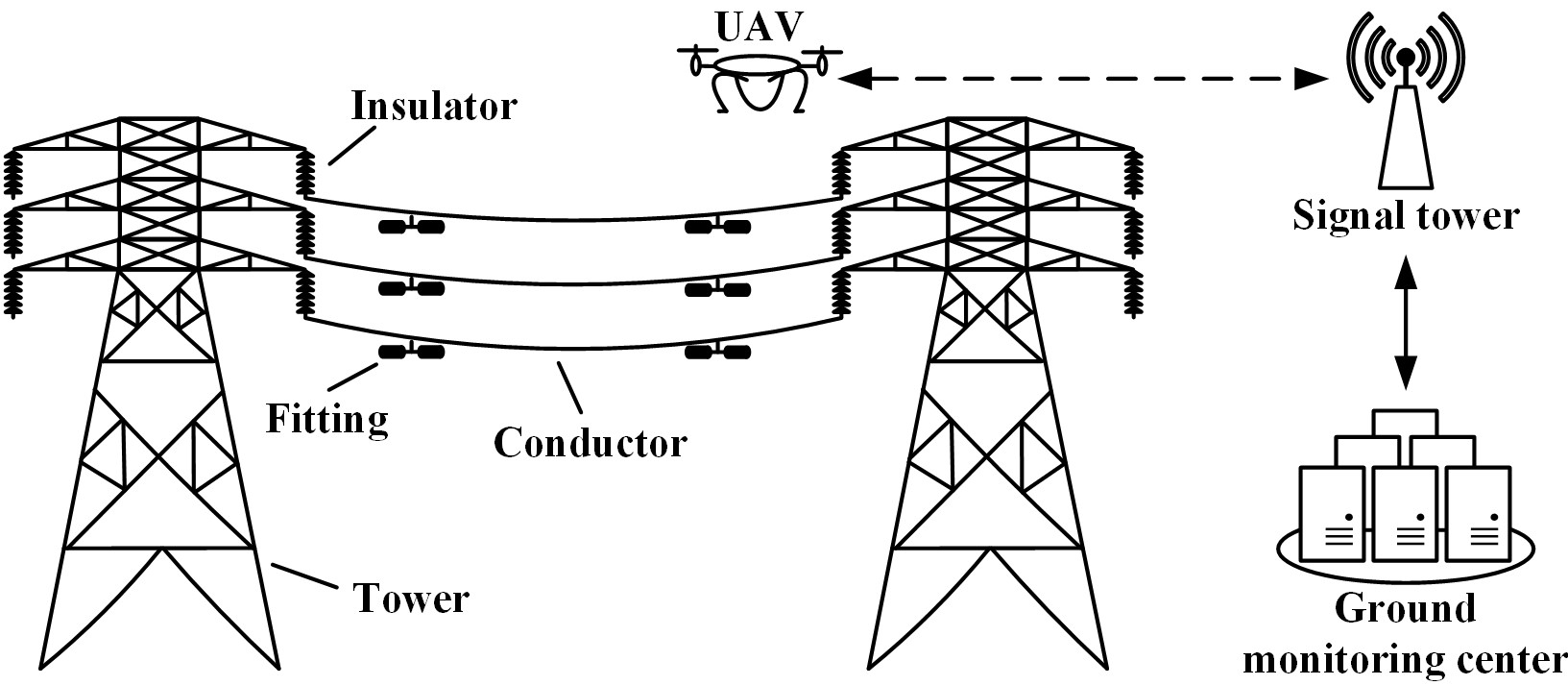}
	\caption{Basic inspection system in power lines.}\label{fig:inspection_system}
\end{figure}

The purpose of component inspection is to identify the condition of power lines and use it as the basis for maintenance decision-making.
\mbox{Fig. \ref{fig:inspection_system}} depicts a fundamental power lines inspection system based on UAVs.
The UAV captures the images of power line components and then sends to the ground monitoring center by wireless communication for further analysis.

\begin{figure}[ht]
	\centering
	\includegraphics[width=8.5cm]{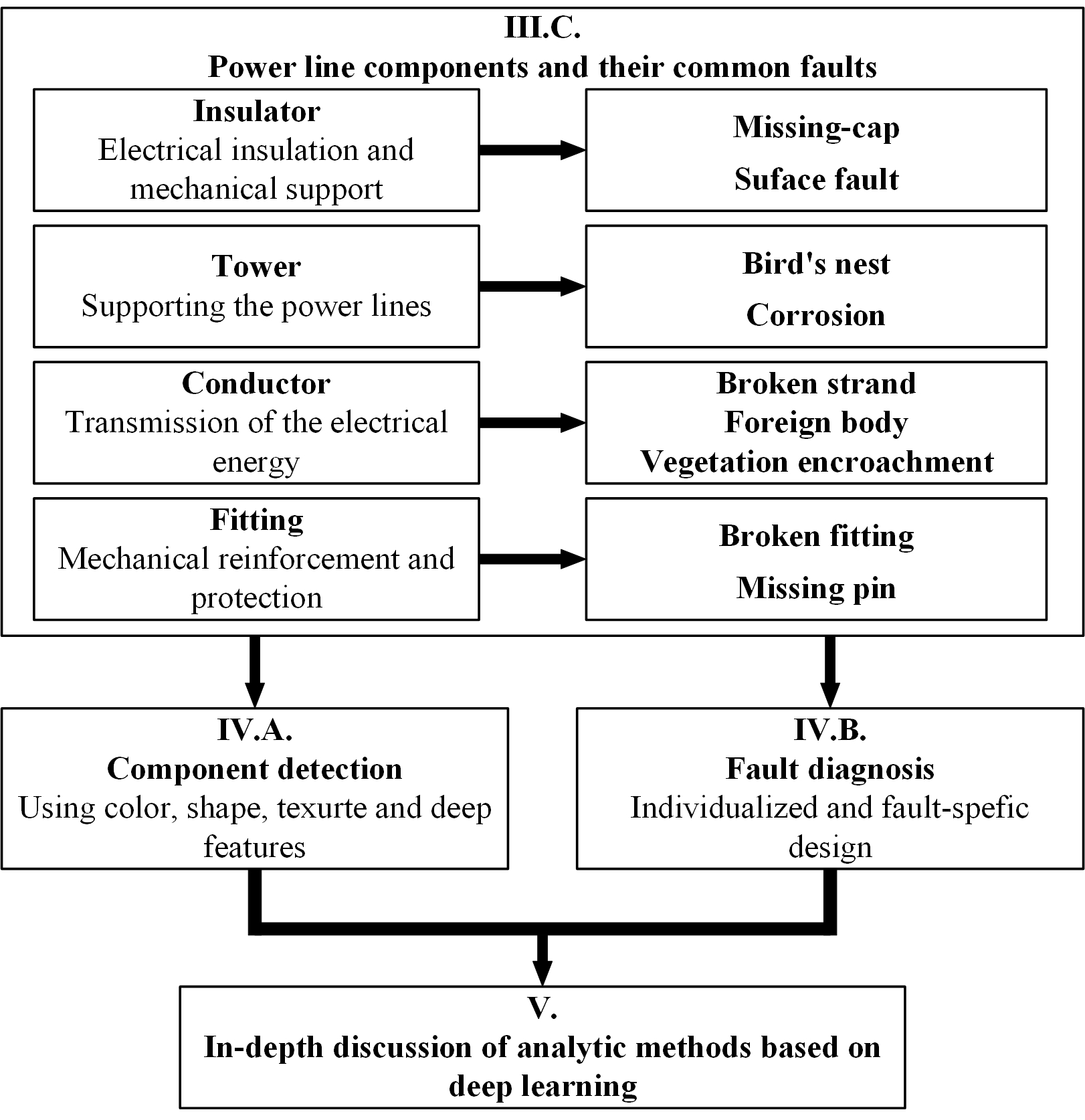}
	\caption{Taxonomy for inspection data analysis in power lines.}\label{fig:taxonomy}
\end{figure}

According to the content of captured aerial images, the main inspection items cam be taxonomically classified into four categories: insulator, tower, conductor and fitting.
In addition, each kind of component has several common fault types.
The detailed taxonomy of the researches reviewed in this paper is illustrated in \mbox{Fig. \ref{fig:taxonomy}}.
The analysis methods of inspection images are classified into component detection and fault diagnosis in the light of their research objective.
Detection of power line components belongs to the object detection task.
In this kind of research, several image features including color, shape, texture and deep features are utilized to locate and classify the component.
Another kind focuses on the diagnosis of faults belonging to components.
Due to the diversity and data scarcity of component faults, the fault diagnosis methods are quite different for different faults in aspects of analytic procedure, applied approach and research popularity.
Therefore, the review of such studies is fault-specific.
Finally, according to these analysis methods, an in-depth discussion of the deep-learning-related researches is provided.

\section{A brief introduction of power lines inspection}
In this section, we first introduce the typical inspection methods (the way to inspect power lines), special attention is paid to the UAV inspection.
Then, we summarize the data source that has been applied in power lines inspection and point out the reasons why visible images are the most widely used.
Finally, we survey the main components and their common faults in power lines while highlight their function, appearance, and potential fault causes.

\subsection{Inspection method}
Conventional power lines inspection methods involve ground inspection and airspace inspection.
Both methods typically identify the condition of power liens by using visual observation \cite{liu2019ground_aerial_inspection}.
The ground inspection is conducted by a team traveling along the power line corridor on foot or by off-road vehicle \cite{aracil2002telerobotic}.
During this procedure, inspectors visually inspect the power lines by using observation tools such as binoculars, infrared cameras and corona detection cameras.
Although the ground inspection has been widely applied for decades due to the high accuracy, but the problems including labour-intensive, low efficiency, and extremely complex landform and weather, all make the ground inspection is gradually replaced by airspace inspection.

The airspace inspection is typically performed by a climbing system or an aerial system.
The former applies a mobile robot to cross obstacles found on power lines and inspects the passing components along the line \cite{fan2018climbing_inspection}.
Climbing robots can obtain high quality images due to its proximity to the conductors.
However, the disadvantages of the climbing system limit its application including the damage to lines, low efficiency, incomplete inspection, and obstruction by obstacles.
The aerial system inspects power lines based on aerial vehicles such as helicopter, multi-rotor UAV and fixed wing UAV \cite{deng2014aerial_platform}.
The aerial vehicle travels along the power line which is controlled by human or flies automatically.
During this procedure, multiple sensors on-board the aerial vehicle are utilized for visual observation and data collection.
Several advantages of aerial system make it a routine inspection method:
1) Access to hard-to-reach locations which means the high flexibility in data acquisition.
2) Capable of loading multiple sensing devices for inspection.
3) Address the problems of low efficiency and damage to lines.

Among the aerial inspection system, the multi-rotor UAV inspection offers a further level of superiority over other inspection methods \cite{menendez2019position}.
The reasons are as follows:
The multi-rotor UAV can fly relatively close to power lines to capture detailed images of power line components.
In addition, it is much cheaper than other aerial vehicles with low operation cost.
Therefore, power lines inspection based on multi-rotor UAV has become the mainstream inspection method.

\subsection{Data sources}
The inspection data acquired from different inspection methods should be analyzed by human or computers to identify the condition of power lines.
Different types of data (or different data sources) have different data analysis methods.
Hence, it is important to determine the data source in a power lines inspection system.
In this paper, we summarize the data sources into two main categories: image data and non-image data.
The non-image data mainly refers to the airborne laser scanner (ALS) data which is also known as georeferenced point cloud data \cite{chen2018LiDAR_inspection}.
It can generate detailed 3D data with the coordinate information of objects and has been applied in the mapping and 3D reconstruction of the power line corridor.
Besides that, the text data such as inspection information and flight record also belongs to non-image but there are rare practical applications based on that.

The image data is the major data source in the of power lines inspection because most conditions of the power lines can be identified through visual observation.
The image data mainly includes visible images \cite{jenssen2019ssd_multi}, infrared images \cite{zhao2016infrard_ins}, ultraviolet images \cite{chen2019ultraviolet_ring}, synthetic aperture radar images \cite{wang2019radar_inspection}, and optical satellite images\cite{michalski2019satellite_tower}.
The infrared image reflects the temperature of objects that can be applied to detect the abnormal heat.
The ultraviolet image is typically used to detect corona discharges of the power lines.
The synthetic aperture radar image and optical satellite image provide large-area coverage that have been used in vegetation monitoring near the power lines.
Among data sources belonging to image data, the visible image is the most widely used data source in power lines inspection due to the following advantages:
1) The vast majority of the faults have visible characteristics and can be diagnosed by visible visual observation.
2) The visible image is more appropriate to the intuitive habit of human.
3) The acquisition of visible images is flexible, low-cost and high-quality that benefited from the well-developed visible camera and aerial photography technology.

\subsection{Power line components and their common faults}
The inspection of power line components is the fundamental task and is among the most popular research topic in the field of power lines inspection.
The objective of this task is to identify the condition of these components and check for faults that should be maintained.
There are many types of components including tower, conductor and accessories (e.g., insulator and fitting) attached to them, and each component type has various faults.
In this paper, we summarize the power line components into four categories including insulator, tower, conductor and fitting \cite{lan2018rcnn_multi}.

\subsubsection{Insulator}
The insulator is an essential component with the dual function of electrical insulation and mechanical support in power lines.
As can be seen in \mbox{Fig. \ref{fig:component_samples} (a)}, the insulator has a repetitive geometric structure with stacked caps.
Depending on the voltage level and nearby environment of the power line, the appearance of insulators is different in color, size and string number (e.g., single string and double strings).
Due to the outdoor working environment, insulators are exposed to the weather especially thunder-strike and icing which can make them malfunction.
The common faults of insulators are missing-cap and surface fault.
The missing-cap refers to one or more caps falling off the insulator that can be seen in \mbox{Fig. \ref{fig:fault_samples} (a)}. 
The surface fault would reduce the insulation ability that occurs to the surface of insulator cap including flashover (see \mbox{Fig. \ref{fig:fault_samples} (b)}), icing and pollution.

\subsubsection{Tower}
The role of towers is to support power lines for maintaining the safety distance between conductors and the ground.
There are two forms of tower appearance: lattice-like structure and pole-like structure that can be seen in \mbox{Fig. \ref{fig:component_samples} (b)}.
Generally, the former is made of lattice steel with metallic surface while the later is constructed by reinforced concrete.
Two common faults of towers that should be taken into considered in the inspection are corrosion and bird's nest.
As can be seen in \mbox{Fig. \ref{fig:fault_samples} (c)}, the corrosion (also known as deterioration) occurs on the surface of tower materials that would shorten the service life of towers.
Bird encroachment is another tower fault threatening the safety of power liens which can be seen in \mbox{Fig. \ref{fig:fault_samples} (d)}.
Birds nesting on towers would affect the tower's insulation performance and cause trip accident.

\subsubsection{Conductor}
Conductors are generally made of copper or aluminum that are utilized to transport the electrical energy.
Depending on the photography distance, the conductor has different appearances in the aerial image which is shown in \mbox{Fig. \ref{fig:component_samples} (c)}.
In the long distance, the conductors can be treated as slender parallel lines.
When the camera is close to conductors, they present the appearance of spiral strips.
The conductor faults that cause frequently are vegetation encroachment, broken strand and foreign body.
The power lines cover a wide area and sometimes cross the forests that nearby growing trees may touch the conductor and then cause short-circuit accidents.
An example of vegetation encroachment is shown in \mbox{Fig. \ref{fig:fault_samples} (e)}
The broken strand is generally caused by conductor galloping and heating that can be seen in \mbox{Fig. \ref{fig:fault_samples} (f)}.
The foreign body such as kite, ballon and plastic bag hanging on conductors by the wind would threaten the safety of power system.

\subsubsection{Fitting}
\begin{figure}[ht]
	\centering
	\includegraphics[width=8.5cm]{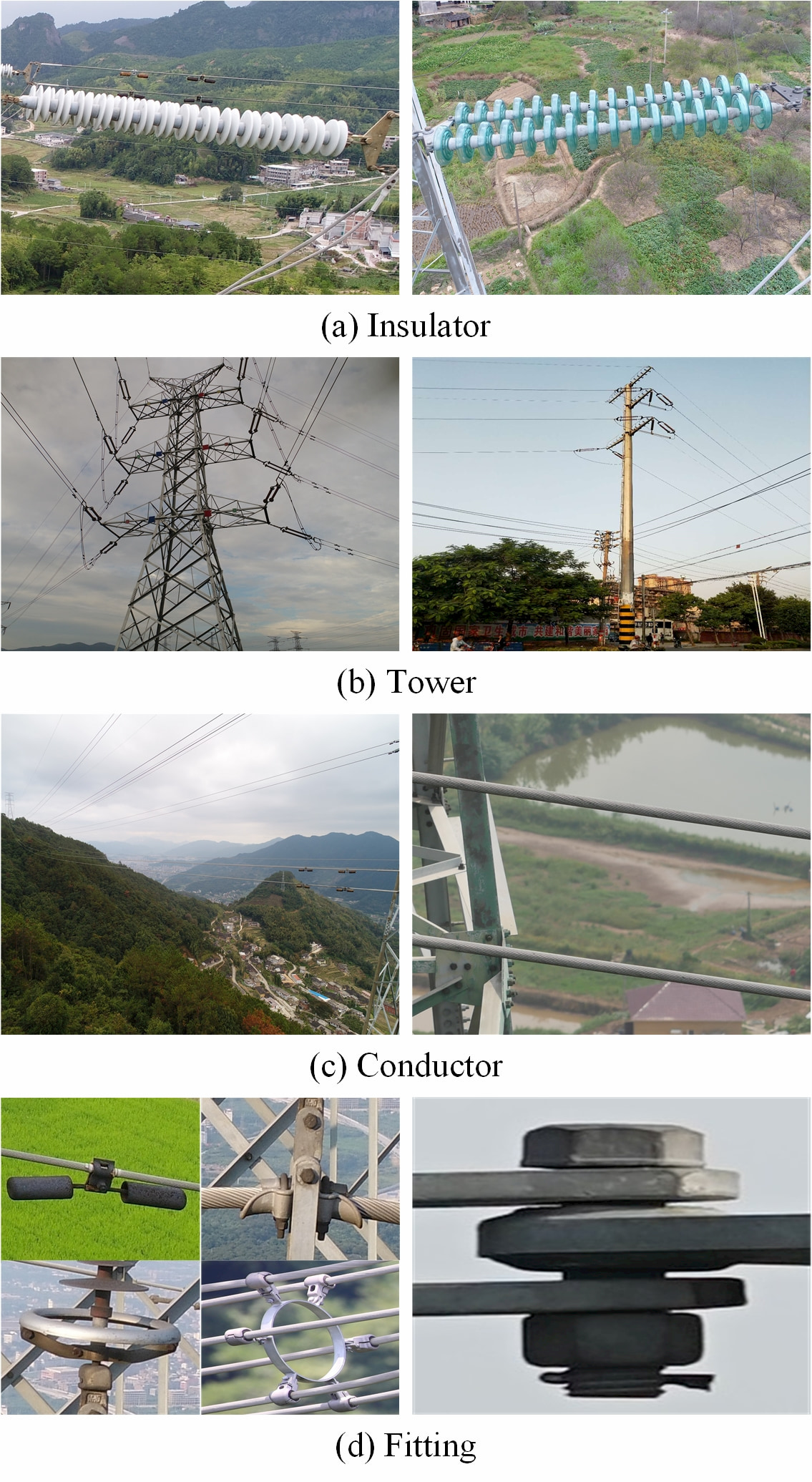}
	\caption{Samples of the power line component}\label{fig:component_samples}
\end{figure}

\begin{figure}[ht]
	\centering
	\includegraphics[width=8.5cm]{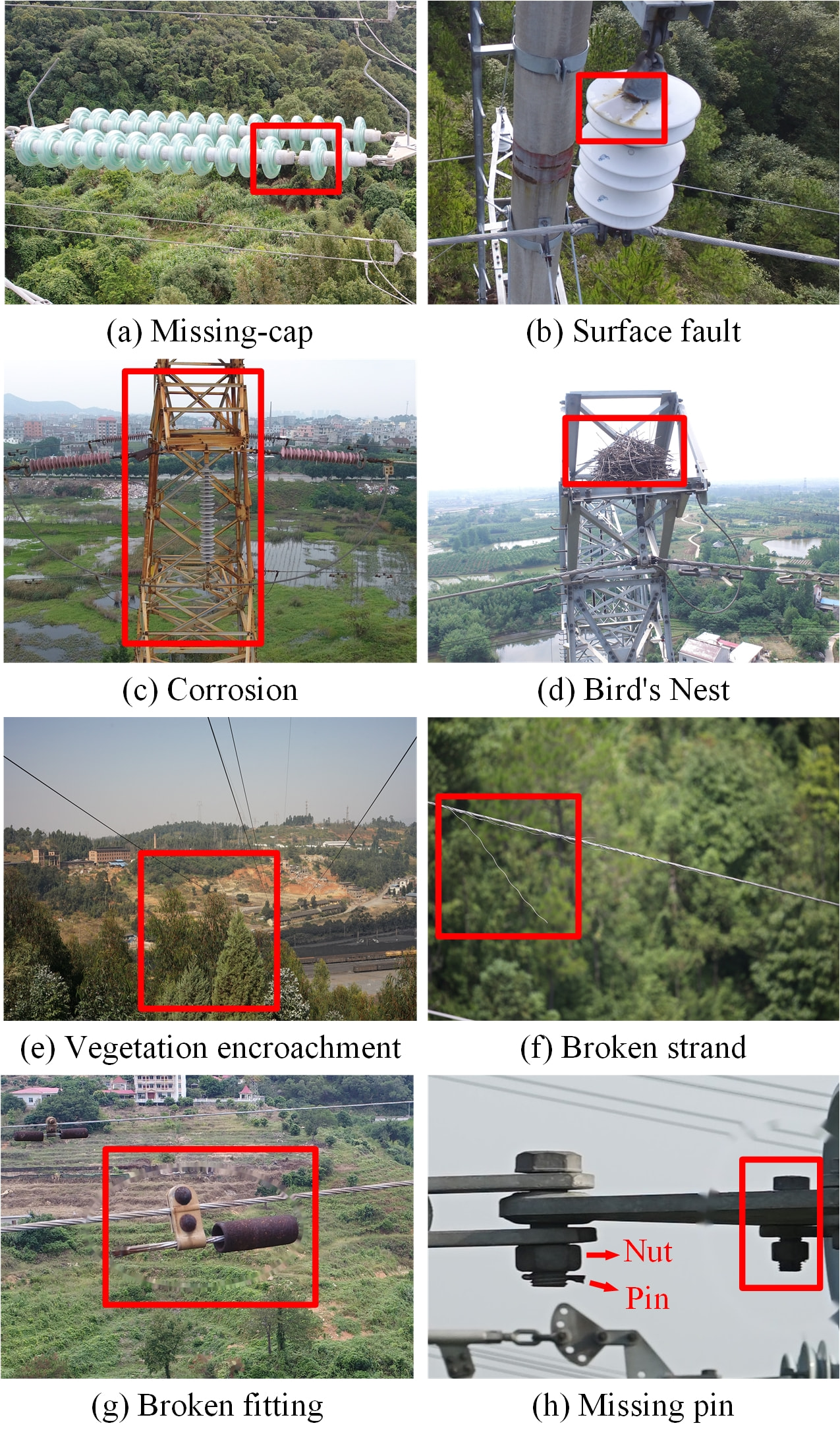}
	\caption{Samples of the common fault in power lines}\label{fig:fault_samples}
\end{figure}

The role of fittings is to reinforce and protect other components such as insulators and conductors.
Due to the variety of fittings, the category of fitting has some subclasses including damper, clamp, arcing ring, spacer , and fastener that are shown in \mbox{Table. \ref{fig:component_samples} (d)}.
With the increasing service life of power lines, part of fittings became invalid causing other components to loosen or even fall off.
Broken fitting is a common fault that fittings show signs of corrosion, wear, cracking, and loosening.
\mbox{Fig. \ref{fig:fault_samples} (g)} shows a broken damper with missing half of the body.
Fasteners are widely used fittings in power lines for mechanical reinforcement which are composed of bolt, nut and pin.
Missing pin is another common fault of fittings which can be seen in \mbox{Fig. \ref{fig:fault_samples} (f)}.
The left is the normal fastener while the right fastener in the red bounding box lost its pin.

\section{Literature review of data analysis in power lines inspection}
In this section, the works on inspection data (almost visible images) analysis are reviewed from two
perspectives.
The first is component detection.
It is very important not only for further fault identification, but also can be used in other practical applications such as UAV navigation, resource management, and video tracking.
Researches about component detection are divided into five groups according to the image features they used: color, shape, texture, fusion and deep.
The second is fault diagnosis which is equally important for determining the condition of power lines.
The works on fault diagnosis are summarized from the perspective of different fault types including surface fault of insulator, missing-cap of insulator, corrosion of tower, bird's nest, broken strand of conductor, foreign body, vegetation encroachment, broken fitting, and missing pin of fitting.
To elaborate the characteristics of the literature reviewed in this section, two tables (\mbox{Table. \ref{tab:summary_obj}} and \mbox{Table. \ref{tab:summary_diag}}) are made which provide the information of the methods, data and performance.
Some valuable details in the researches are also provided such as classifier, image preprocessing approach, and main image features. 
Finally, two main limitations of current literature are introduced including the insufficient research on some components with their faults and the lack of practical engineering.

\subsection{Component detection}
\begin{figure}[ht]
	\centering
	\includegraphics[width=8.5cm]{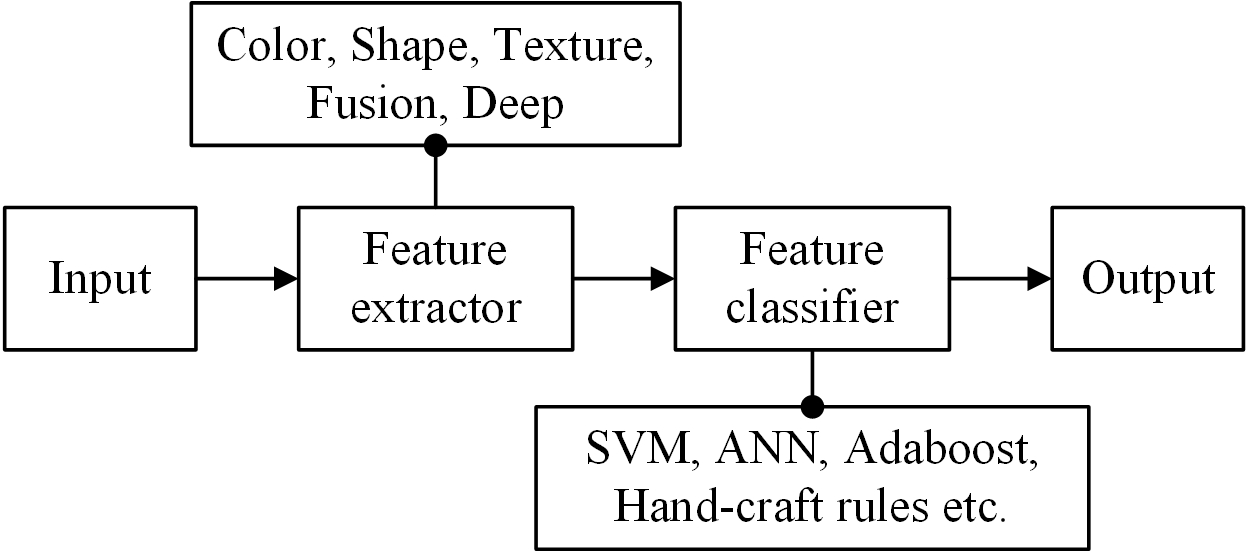}
	\caption{The common procedure of component detection in power lines}\label{fig:det_procedure}
\end{figure}

The detection of power line components is the key prerequisite for further analysis.
The number of research articles dealing with component detection has significantly increased in the last few years.
As can be seen in \mbox{Fig. \ref{fig:det_procedure}}, the common detection procedure can be divided into two stages: feature extraction and feature classification.
Features were extracted from images and then input to the classifier for identifying whether they belong to the component.
In this paper, the extracted features can be grouped into five major categories : color feature, shape feature, texture feature, fusion features and deep feature.
The features beside the deep feature are also defined as hand-craft features or shallow features. 
As for classification stage, the learning-based algorithms are frequently used as the feature classifier such as SVM, ANN, and Adaboost.
Besides that, some hand-craft rules based on the characteristics of power line components are also responsible for classification.
For instance, the insulator has a repetitive geometric structure with multiple caps that have distinctive circular shape.
According to this rule, the insulator can be detected by searching the ellipse in the image.

In the following content, we will summarize the current literature based on different image features with special attention to the core method, component types, image preprocessing approaches, classifier, data for training and testing, and the method performance.

\begin{table*}[ht]  
	\caption{Summary of the related work for components detection.}  
	\label{tab:summary_obj}
	\centering
	\renewcommand\arraystretch{1.2}
	\setlength{\tabcolsep}{2.2mm}{
		\begin{tabular}{ c l l l l m{2.5cm} m{2.5cm}} 
			\hline
			\textbf{Features}  & \textbf{~~~~~~Method} & \textbf{Component} & \textbf{Image preprocessing}  & \textbf{~~Classifier} & \textbf{~~~~~~~~~Data} & \textbf{~~~~~Performance}\\
			\hline
			\multirow{7}{*}{\textbf{Color}}  & Color model \cite{zhang2010simple} & Insulator & \tabincell{l}{RGB to HSI \\ Morphological filter} & Thresholding & Test: 2 & ----\\
			\cline{2-7}
			& Color model \cite{yao2012Color_ins} & Insulator & \tabincell{l}{RGB to HSI \\ Morphological filter} & Rules & Test: 50 & Complete: 50, incomplete: 42\\	
			\cline{2-7}
			& Color model \cite{reddy2013condition}  & Insulator & \tabincell{l}{RGB to Lab \\ K-means cluster} &  SVM & Test: 33 & Recall: 100\% \\
			\cline{2-7}
			& Color model \cite{castellucci2013Color_ANN_tower} & Tower & \tabincell{l}{RGB to HSI \\ RGB to YCbCr} & ANN & Train: 350, Test:350 & Hit rate: 70\% \\	
			\hline
			\multirow{7}{*}{\textbf{Shape}}  & OAD-BSPK \cite{zhao2015OAD-BSPK} & Insulator & \tabincell{l}{RGB to Gray \\ Morphological filter} & Rules & Test:4 & Positioning accuracy:  58.4\%  \\
			\cline{2-7}
			& Canny \cite{tragulnuch2018OBIC_tower} & Tower & \tabincell{l}{RGB to Gray \\ Gaussian filter} & Rules & Test: 2 videos with 25 FPS & Recall: 100\% \\
			\cline{2-7}
			& PLineD \cite{Santos2017PLineD_line} & Conductor & RGB to Gray  & Rules & Test: 82 & ----\\ 
			\cline{2-7}
			& MLP \cite{liu2017MLP_fitting} & Fitting & ---- & Rules & Test: 2000 &  Correct recognition rate: 80.42\% \\
			\cline{2-7}
			& \tabincell{l}{Profile projection\\ + SVM \cite{li2012PP_SVM_ins}} & Insulator & \tabincell{l}{RGB to HSI \\ Morphological filter} & SVM & Test: 637 & Correct rate: 95.01\% \\			
			\hline
			\multirow{6}{*}{\textbf{Texture}}  & GLCM-GMACM \cite{wu2012GMAC} & Insulator & RGB to Gray & K-means & Test: 100 & False alarm rate: 5\% \\
			\cline{2-7}
			& LDP+SVM \cite{jabid2016rotation} & Insulator & ---- & SVM & Test: 325 & Recall: 94.24\% \\
			\cline{2-7}
			& RI-LDP+SVM \cite{jabid2018RI-LDP_SVM_ins} & Insulator & ---- & SVM & Test: 395 & Recall: 95.74\% \\
			\cline{2-7}
			& Harr+AdaBoost \cite{Jin2012Haar_damper} & Fitting & \tabincell{l}{RGB to Gray \\ Smoothing filter} & AdaBoost & Train: 4517, test: 100& True positive rate: 92.48\% \\
			\cline{2-7}
			& HM-LA \cite{Fu2017Haar_fitting} & Fitting & RGB to Gray  & AdaBoost & Test: 21 & Detection rate: 90\% \\
			\hline
			\multirow{4}{*}{\textbf{Fusion}}  & HOG-LBP+SVM \cite{tiantian2017HOG_LBP_ins} & Insulator & \tabincell{l}{Otsu thresholding \\Morphological filter} & SVM & Test: 500 & Right detection rate: 89.1\%  \\
			\cline{2-7}
			& CGT-LBP-HSV \cite{wang2016CGT_LBP_HSV_ins} & Insulator & ---- & Rules & Test: 100 & Recall: 88.9\% \\
			\cline{2-7}
			& ACF+Boost \cite{han2016ACF_tower} & Tower & ----  & Boost & Train: 600, test: 400 & Test error: 3.25\% \\
			\hline
			\multirow{12}{*}{\textbf{Deep}} & CNN+SW \cite{liu2016CNN_softmax_SW_ins} & Insulator & \tabincell{l}{Augmentation\\Resize} & Softmax & Train: 3000, test: 341 & True positive rate: 90.9\% \\
			\cline{2-7}
			& Faster R-CNN \cite{liu2018frcnn_ins} & Insulator  & \tabincell{l}{Augmentation\\Resize}  & Softmax & Train: 3000, test: 1500 & Recall: 87.53\% \\
			\cline{2-7}
			& Faster R-CNN \cite{Wang2017RCNN_fitting} & Fitting  & Resize & Softmax & Train: 4500, test: 1500 & Recall: 84.03\% \\
			\cline{2-7}
			& SSD \cite{xu2018SSD_for_ins}  & Insulator &\tabincell{l}{Augmentation\\Resize} &  Softmax & Train: 2000, test: 500 & Mean average precision: 94.7\% \\
			\cline{2-7}
			& YOLOv2 \cite{wang2018YOLO_for_ins} & Insulator & \tabincell{l}{RGB to Gray\\Resize} & Softmax & Train: 800, test: 200 & Recognition accuracy: 83.5\% \\
			\cline{2-7}
			& YOLOv3 \cite{chen2019YOLO_tower} & Tower & \tabincell{l}{Augmentation\\Resize}  & Logistic & Train: 11951, test: 1478(mixing with simulated and actual images) & Mean average precision: 90.45\% \\
			\cline{2-7}
			& FCNs \cite{hui2018Fst_FCNs} & Conductor  & ---- & Softmax & Train: 400, test: 200 & Accumulative pixel errors : 450 pixels \\
			\cline{2-7}
			& cGAN \cite{chang2018cGAN_for_line} & Conductor  & \tabincell{l}{Augmentation\\Resize} & Discriminator & Train: 5000, test: 1000 & Accuracy rate: 94.8\% \\
			\hline 
		\end{tabular}  
	}
\end{table*}

\subsubsection{Color feature}
Detection of power line components has been investigated in few studies related to color feature.
In all studies, the images were converted to a specific color space and most of the studies concentrated on HSI(Hue, saturation, intensity) color space.
Zhang et al. \cite{zhang2010simple} obtained the intensity image by converting the aerial image into HSI color space from RGB color space.
Then, the morphological filter is utilized to denoise, and the connects components analysis is proposed to locate the possible area of insulators.
Finally, the glass insulator is detected through screening these areas by color thresholding. 
Some images describing the detection process are used as the results the research.
Yao et al. \cite{yao2012Color_ins} also converted the aerial image into HSI color space and the saturation image was used to recognize insulators.
The morphological filter and Optimal Entropic Threshold (OET) were applied for contour extraction.
Contours belonging to insulators were identified according to the factors in hand-craft rules (e.g., circularity, duty-factor, Hu-moment Invariant).
The method was tested in 50 inspection images.
They found that all the complete insulators were correctly detected while 8 incomplete insulators were miss detected.

Some studies concentrated on the Lab color space for insulator detection.
Reddy et al.\cite{reddy2011dost,reddy2013condition} converted the RGB　image to Lab color space and obtained the required cluster by applying K-Means.
The potential bounding box that may contain the insulator was drew by thresholding.
Then, the color feature of each candidate box was fed into the trained ANFIS \cite{reddy2011dost} or SVM \cite{reddy2013condition} for identifying the correct box.

The combination of different color spaces was also discussed by Castellucci et al. \cite{castellucci2013Color_ANN_tower}.
They investigated the tower detection approach based on color features of HSI and YCbCr(Luma, blue-difference, red-difference) color space.
Color maps were obtained by converting the aerial images into HSI and YCbCr color space respectively.
Then, Channels B, S and Cr from these color maps were utilized to compose the input vector of the ANN.
The 3-layer ANN classified the color features into four class: pole, crossarm, vegetation and others.
In this research, a transmission tower consisted of a pole and a crossarm.
Therefore, the tower can be detected once the pole and crossarm are found.
Totally 700 images were utilized in this research and the hit rate of 70\% was achieved.

To summarize, the color feature represents the global information more than the local information, which limits its practical application.
Further, how to determine the range of color values is a challenging problems in the complex background of power lines.
Hence, most of the studies based on the color feature are early researches (before the year of 2013) in the field of power lines inspection.

\subsubsection{Shape feature}	
Compared to the color feature, the shape feature shows better representation of power line components due to their line-based structure.
In most studies, the contours or edges were extracted for further classification by using sharpening edge \cite{zhao2015OAD-BSPK}, Canny edge detector \cite{tragulnuch2018OBIC_tower}, edge drawing \cite{Santos2017PLineD_line} and crossing gradient template \cite{liu2017MLP_fitting}
.

Zhao et al. \cite{zhao2015OAD-BSPK} proposed an insulator detection method based on Orientation Angle Detection and Binary Shape Prior Knowledge (OAD-BSPK).
During image preprocessing, the binarization and morphological filter were performed to obtain binary image.
Then, the orientation angle was computed by using sharpening edge, and was used to rotate the binary image that made insulator vertically.
According to the binary shape prior knowledge of insulators and the possible orientation angles, small regions were removed thus the insulator was detected.
Four real-world aerial images were used to evaluated the proposed method.

In stead of sharpening edge, Tragulnuch et al. \cite{tragulnuch2018OBIC_tower} detected power towers based on a commonly used edge detector called Canny.
At first, Canny edge detector was utilized to extract the contours.
Then, the image was separated into 10 $\times$ 10 pixel boxes and Hough line transformations was applied to obtain straight-line.
The box that have long straight-line pass through it was marked as the candidate box.
Finally, the hand-craft rules such as the length and number of the straight-line were used to remove the false box and classify the power tower.
The method was tested in two inspection videos that have 1920$\times$1080 pixels resolution with 25 frames per second.
Results showed that all the towers appeared in videos were correctly detected.

By using Edge Drawing, Santos et al. \cite{Santos2017PLineD_line} studied the detection of power conductors.
First, Straight line segments were extracted through Edge Drawing.
Then, the hand-craft rules consisted of four steps were designed to identify these segments.
Step 1 was cutting the bending segments into horizontal segments and vertical segments.
In step 2, the short segments were removed according to the covariance between each segment.
The rest segments were grouped on the basis of line spacing in the step 3.
Finally in step 4, the segments belonging to the conductor were picked out according to the number of parallel lines in each group.
In the experiment, they extracted all the conductors in 82 real-world aerial images.

The crossing gradient template was applied for damper detection in the research of Liu et al. \cite{liu2017MLP_fitting}.
The detection scheme so-called multi-level perception consisted of three perception levels including low-level, middle-level and high-level.
The low-level perception adopted crossing gradient template for segments extraction.
In the middle-level perception, the aerial image was firstly divided into multiple blocks, and then the parallel lines and cross lines were utilized to define the conductor area and the tower area respectively.
Finally in the high-level perception, the power line components were recognized according to the designed hand-craft rules.
The rules were based on the local contour feature of damper and position relation between damper, tower and conductor.
The algorithm was evaluated at real-world images that 1608 dampers were correctly detected among the whole 2000 dampers in the dataset. 

The aforementioned researches utilized hand-craft rule as the classifier.
The reasons account for this phenomenon were as follow: the power line components such as towers and conductors have obvious linear structure compared to the background in the aerial images.
Once the shape feature such as contours and edges were obtained, we can design some simple rules, for example, the length, number or positional relationship of the segments, to filter the extracted shape features.
Then, the components can be detected after several filtering operations.
However, in addition to the segment itself, some deeper information of the shape feature was worth studying, and the learning-based method is another good choice for feature classification.
Li et al. \cite{li2012PP_SVM_ins} provided an example who introduced a profile projection method to locate the potential area of insulators.
Next, the principal component analysis was introduced for tilt correction of the potential area.
After that, shape feature was derived from vertical profile projection curve.
Finally, the trained SVM was utilized to indicate the extracted features of insulators.
In the experiments, 637 cropped images were used to test the proposed method, and correct rate of 95.01\% was obtained. 

\subsubsection{Texture feature}	
The following studies discussed the detection of power line components based on texture feature and most of them concentrated on insulators \cite{wu2012GMAC,jabid2016rotation,jabid2018RI-LDP_SVM_ins} and fittings \cite{Jin2012Haar_damper,Fu2017Haar_fitting}.
Contrast to the color feature, the texture feature more characterize the local feature that was appropriate for the detection of those components with repetitive geometric structure (e.g., insulator, damper, and spacer).

Wu et al. \cite{wu2012GMAC} introduced texture segmentation algorithm for insulator detection.
The texture feature was extracted by Gray Level Co-occurrence Matrix (GLCM) and classified into two classes by K-means. 
Then, insulators were recognized by means of the Global Minimization Active Contour Model (GMACM).
Experiments on 100 aerial images with 5\% false alarm rate demonstrated the performance of the proposed algorithm.
Local Directional Pattern (LDP) was a commonly used method for texture feature extraction and applied in some studies for insulator detection.
Jabid et al. \cite{jabid2016rotation} dealt with the orientation variation problem in the insulator detection.
The proposed method presented in the article consists of three steps: correcting the orientation of insulators into horizontal, performing LDP to extract texture feature, and classifying the texture feature based on SVM.
They established a evaluation set contained 325 images to verify the presented algorithm and achieved the recall rate of 94.24\%.
In later research \cite{jabid2018RI-LDP_SVM_ins}, they improve the LDP method to solve the issue of orientation variation which called Rotation Invariant LDP (RI-LDP).
Thus, the step 1 of detection scheme in \cite{jabid2016rotation} which needs to correct the insulator orientation can be removed.
The SVM still applied as the feature classifier.
The evaluation set increased to 395 image with 722 labeled insulators and this improved method achieved 95.74\% recall.

Besides insulators, there are some studies focused on the fitting detection based on Haar-like features.
Jin et al. \cite{Jin2012Haar_damper} extracted Haar-like features to detect dampers.
The cascade Adaboost classifier was used to identify the features from sliding windows of original image.
Totally 4517 images with 1518 damper images and 2999 background images were collected for training the classifier and 100 images were used for testing.
Results showed the effectiveness of the proposed method with 92.48\% true positive rate.
Fu et al. \cite{Fu2017Haar_fitting} also concentrated on the detection of fittings such as dampers and fasteners.
In stead of detecting the entire component, they decomposed it into multiple sub components and detected them respectively.
The combination of the Haar-like feature and AdaBoost classifier were used for recognition of these sub components.
Then, the damper or nut can be detected according to the positional relationship of the sub components.
The method was evaluated at 21 images and achieved over 90\% detection rate under simplex photography situation.

\subsubsection{Fusion Feature}	
A few attempts have been made to detect power line components based on fusion features.
In the following studies, multiple types of features (e.g., shape, color, and texture) were combined for components detection.
Yan et al. \cite{tiantian2017HOG_LBP_ins} discussed the use of fusion feature for insulator detection.
The Histogram of Oriented Gradients (HOG) and Local Binary Pattern (LBP) features were extracted and then classified by SVM.
The SVM classifier was trained with 700 local sub insulator images from aerial videos.
The proposed method was evaluated at 500 images with 89.1\% detection rate.
Authors also discussed the benefit of the fusion feature compared to single feature method.
The HOG-based method and LBP-based method achieved 85.1\% and 81.8\% detection rate separately.
The results illustrated that the fusion feature showed more capacity for the representation of insulators.
Authors in \cite{tiantian2017HOG_LBP_ins} mentioned that the fusion feature can achieve higher accuracy than the single feature.
Wang et al. \cite{wang2016CGT_LBP_HSV_ins} proposed an insulator detection method that merged the shape, color and texture features.
As for shape feature, the edges were extracted using different directions gradient operators.
Then the candidate regions were produced by parallel lines clustering.
With respect to color and texture features, HSV color space converting and LBP were performed on the candidate regions.
Finally, the insulator can be detected by similarity calculation based on the Euclidean distance of HSV and LBP features.
Experiments were implemented on 100 images and 88.9\% detection rate was achieved.

The methods mentioned above classified different features separately, the following study polymerized different features into a multi-channel feature map for classification. 
Han et al. \cite{han2016ACF_tower} described a process for tower detection based on the fusion feature in 10 channels.
The Aggregate Channel Features (ACF) computed several feature channels including 1 channel of normalized gradient magnitude, 6 channels of histogram of oriented gradients and 3 channels of LUV color space.
After the feature extraction, the Adaboost classifier was utilized to distinguish towers from background.
The proposed method was tested by using 200 images and attained 96.75\% accuracy.

Although the application of fusion feature for power line component detection is rare, it still shows considerable potential under the situation of data insufficiency.
Compared with single feature methods, fusion features can describe the components more comprehensively, which means higher accuracy can be obtained.
However, this improvement was based on the sacrifice of detection speed due to the extraction of multiple features.

\subsubsection{Deep Feature}
The number of research articles dealing with component detection of power lines based on deep learning has significantly increased in the last few years, especially since 2016.
Theses researches extracted deep feature from aerial images for component detection, and most of them achieved better performance than the researches based on hand-craft features that mentioned above.
The comparative experiments can be found in papers \cite{liu2016CNN_softmax_SW_ins,hui2018Fst_FCNs,chang2018cGAN_for_line}.
In deep learning approaches, the data quantity is an important factor for their performance.
Thus, data augmentation was applied in order to solve data insufficiency in researches \cite{liu2016CNN_softmax_SW_ins,liu2018frcnn_ins,xu2018SSD_for_ins,chen2019YOLO_tower,chang2018cGAN_for_line}.
Resizing of the images also became a common process that mentioned in \cite{liu2016CNN_softmax_SW_ins,liu2018frcnn_ins,Wang2017RCNN_fitting,xu2018SSD_for_ins,wang2018YOLO_for_ins,chen2019YOLO_tower,chang2018cGAN_for_line}.
There are two main reasons for resizing: on the one hand, some deep learning frameworks required fixed size input;
On the other hand, aerial images collected from UAV had high resolution.
Resize the image to a smaller size can save a lot computation resource. 

In the early research of component detection based on the deep feature, the simple CNN combined with sliding window was introduced.
Liu et al. \cite{liu2016CNN_softmax_SW_ins} introduced a deep-learning-based method for insulator recognition.
A six-layer convolutional neural network combined with sliding windows scheme was applied for the detection of insulators.
They evaluated the method by using 341 images and achieved 90.9\% true positive rate.
The comparative experiments were also conducted with Bag of word (Bow) and Deformable Parts Model (DPM with HOG feature), the result demonstrated the improvement of the proposed method compared to these shallow-feature-based methods.

With the development of deep learning technology, a large number of famous object detection frameworks have emerged in recent years.
Researchers in the field of power line inspection attempted to introduce these existing frameworks into the detection of components.
For example, Liu et al. \cite{liu2018frcnn_ins} applied Faster Regions with Convolutional Neuron Network (Faster R-CNN) to detect insulators in the aerial image.
Wang et al. \cite{Wang2017RCNN_fitting} also employed Faster R-CNN for fitting detection including dampers, spacers and arcing ring.
These two researches both cropped the aerial image with object as main part in the center and then resized this sub-window to 500 $\times$ 500 resolution.
The insulator detection was also investigated by using Single Shot multi-box Detector (SSD) in the paper of Xu et al.\cite{xu2018SSD_for_ins}, and You Only Look Once v2 (YOLOv2) in the article of Wang et al. \cite{wang2018YOLO_for_ins}.
Pixel sizes of the aerial image were resized to 512$\times$512 for SSD and 448$\times$448 for YOLOv2.
As for tower detection, Chen et al. \cite{chen2019YOLO_tower} trained five YOLOv3 models with various pixel sizes containing 288$\times$288, 352$\times$352, 416$\times$416, 480$\times$480 and 544$\times$544.
Due to the lack of real-world inspection data, they generated 13,429 simulated images for training and testing.
The results showed that the model trained with 352$\times$352 pixel size can achieve 90.45\% mean Average Precision (mAP).

The process to detect conductors based on deep feature is quite different from other components.
In stead of region-based framework, the researchers were more inclined to use pixel-wise framework due to the slender line characteristic.
Hui et al. \cite{hui2018Fst_FCNs} employed the Fully Convolutional Networks (FCNs) to detect transmission conductors from aerial images.
A sequence of images collected from aerial videos were utilized to evaluate the proposed method.
Results showed the improvement of the deep-feature-based method compared with edge-based method.
Chang et al. \cite{chang2018cGAN_for_line} utilized conditional Generative Adversarial Nets (cGANs) to detect the conductor.
For model training, they constructed a specific dataset including four types of conductor images: normal (clear strip texture), linear (slightly farther than the normal ones), quadrangular(emphasize the strip
texture by close observation), noWire (background only).
Meanwhile, data augmentation was applied and the images were all resized to 256$\times$256. 
The proposed method was tested by using 1000 images (500 for simplex samples and 500 for complex samples) and achieved 94.8\% average accuracy.
Comparison experiments were conducted with shallow-feature-based methods such as Line Segment Detector (25.2\%) and HOG (19.4\%), and other deep-feature-based methods such as PCANet (86.8\%) and ENet (95.4\%).
The result illustrated the high efficiency of the deep feature.

In this section, we only introduce several representative works that utilize deep features for component detection.
There are some other researches that apply deep learning method to analyze inspection data, which will be further reviewed in Section V.B.
A detail and in-depth discussion with special attention paid to deep learning is provided.

\subsubsection{Remarks}
\mbox{Table. \ref{tab:summary_obj}} provides the valuable information of researches in power line component detection, which includes the main image features used in the proposed method, inspection component, image preprocessing operation, classifier for the extracted features, brief description of data, and the method performance.

The component detection is a relatively mature area since it has many applications and large available data.
In a majority of existing works, the image feature extractor is manually designed according to the characteristics of components while the feature classification is mainly implemented by the hand-craft rules and shallow learning models.
There are some attempts in applying deep learning models to achieve end-to-end component detection, but the related investigation is still limited.
To improve the performance of component detection, there are at least two ways: 
1) using refined aggregated features instead of single feature.
2) improving deep learning networks based on the characteristics of different components that are distinguished from other generic objects.

For the category of detected component, the insulator has received most of the attention.
To fully monitor the condition of power lines, other component types would need to be further concerned especially the fitting.
In addition, we also find that the description of the experimental data is unclear in part of the literature.
The data quality is an important factor that greatly influences the evaluation of the proposed method.
This information, such as the data size, image resolution, data collection approach and samples for visualization, should be well introduced.
Furthermore, evaluation metrics used in current works are inconsistent.
Many metrics have been applied to illustrate the performance of the proposed method such as recall, precision, accuracy, true positive rate, and average precision.
Even the same metric may have different definitions in different researches.
Besides, we notice that in the existing literature, the authors evaluate the method based on their own private dataset and the comparative experiment is quite limited.
Without the same evaluation metrics and dataset, the superiority of a certain method cannot be guaranteed.
A standard evaluation baseline including metrics and open dataset will promote the research in the whole area of inspection data analysis.


\subsection{Fault diagnosis}
Here, we consider the fault diagnosis of power line components by using visible inspection images.
The fault diagnosis researches are much less than the component detection due to the following reasons:
1) faulty components do harm to the power system, but they are relatively rare compared to normal components. 
2) there are multiple types of faults in the same component. 
3) there are many manifestations of the same fault type in images.
The reasons mentioned above lead to the lack of fault data that limits the use of learning based approaches, while the hand-craft based methods are difficult to deal with such a variety of component faults.

\begin{figure}[ht]
	\centering
	\includegraphics[width=8.5cm]{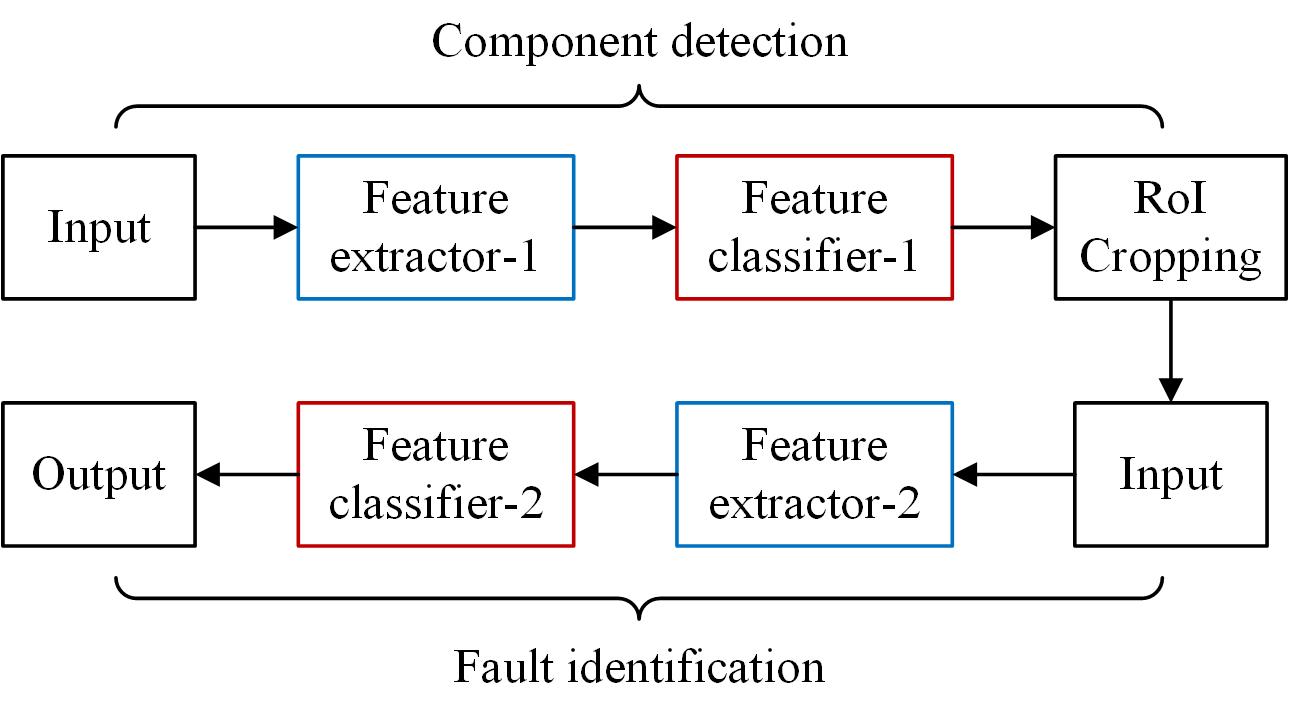}
	\caption{The common procedure of fault diagnosis}\label{fig:dia_procedure}
\end{figure}

As can be seen in \mbox{Fig. \ref{fig:dia_procedure}}, the typical procedure of the fault diagnosis composed of two stages: detecting the component and identifying the fault.
At the first stage, the component region as the Region of interest (RoI) should be detected and cropped in order to filter out background for further analysis.
Then in the second stage, the fault identification method can be applied in the RoI.
Notice that in few studies (e.g., \cite{yang2017IULBP, maeda2017DELM-LRF}), the component detection stage was not considered since the component was already the principal part in the image.
On the other hand, the existence of some objects is a kind of fault such as bird's nest \cite{Xu2017HSV_GLCM_tower,Lu2018CF-CC_tower} and foreign body \cite{Song2015CED-HT_fitting,Tang2018Fst_fitting}, these types of faults are obvious enough to be analyzed directly without the stage of component detection.

In the following content, the literature will be summarized according to the fault categories with special attention to the fault identification stage, while the image features, data, and performance are also concerned.

\begin{table*}[ht]  
	\caption{Summary of the related work of fault diagnosis}  
	\label{tab:summary_diag}
	\centering
	\renewcommand\arraystretch{1.2}
	\setlength{\tabcolsep}{1.8mm}{
		\begin{tabular}{ m{1.7cm} l l m{2cm} l m{2.5cm} m{2.5cm} } 
			\hline
			\textbf{~~~~~Fault} & \textbf{~~~~~~Method}  & \textbf{~~~~Detection} & \textbf{Identification} & \textbf{Main features} & \textbf{~~~~~~~~~Data} & \textbf{~~~~~Performance}\\
			\hline
			\multirow{6}{1.7cm}{Surface fault of insulator} & IULBP \cite{yang2017IULBP} & ---- & IULBP+Rules & Texture & ---- & ----\\
			\cline{2-7}
			& GSS-GSO \cite{Hao2018GSS-GSO_ins_ice} & GrabCut & Rules & Shape & ---- & ----  \\
			\cline{2-7}
			& M-SA \cite{Zhai2018M-SA_ins_flash} & F-PISA & Color model & Color & Test: 100 & Detection rate: 92.7\% \\
			\cline{2-7}
			& CGL-EGL \cite{oberweger2014CGL-EGL_ins}  & CGL & EGL & Shape & Test: 20 instances & True positive rate: 95\%\\
			\cline{2-7}
			& M-PDF \cite{zhao2016M-PDF_ins} & OAD-BSPK \cite{zhao2015OAD-BSPK} & AlexNet & Deep & Train: 300, test: 700 & Mean average precision: 98.71\% \\
			\hline
			\multirow{12}{1.7cm}{Missing-cap of insulator}  & GLCM \cite{wang2016CGT_LBP_HSV_ins} & CGT-LBP-HSV & GLCM+Rules & Texture & ---- & ---- \\
			\cline{2-7}
			& S-AM \cite{zhai2017S-AM_ins_drop}  & Saliency detection & Adaptive morphology & Fusion & Test: 100 & Detection rate: 92.4\% \\
			\cline{2-7}
			& SMF \cite{zhai2018SMF_ins}  & Color model & Morphology & Fusion & Test: 74 & Detection success rate: 91.7\% \\
			\cline{2-7}
			& M-YOLO+AM \cite{Han2019MYOLO_AM_ins}  & M-YOLO & Adaptive morphology & Shape & Test: 42 & Recall: 93.3\% \\
			\cline{2-7}
			& \tabincell{l}{Faster R-CNN \\+ U-net\cite{ling2018Fst-Unet}}& Faster R-CNN & U-net & Deep & Train: 165, test: 55 & Recall: 95.5\% \\
			\cline{2-7}
			& \tabincell{l}{R-FCN \cite{li2018insulator_rfcn}}& ---- & R-FCN & Deep & Train: 2626, test: 500 & Mean average precision: 90.5\% \\
			\cline{2-7}
			& \tabincell{l}{Up-Net+CNN \cite{sampedro2019UpNet_CNN_ins}}& Up-Net & CNN & Deep & Train: 2400, test: 400 (synthetic images) & Accuracy rate: 98.8\% \\
			\hline
			\multirow{3}{1.7cm}{Corrosion of tower}  & DELM-LRF \cite{maeda2017DELM-LRF}  & ---- & DELM-LRF & Deep & Train: 2237, test: 560 & F-measure: 79.6\% \\
			\cline{2-7}
			& CMDELM-LRF \cite{maeda2018CMDELM-LRF}  & ---- & CMDELM-LRF & \tabincell{l}{Deep\\(visual+text)} & Train: 2414, test: 603 & F-measure: 88.8\% \\
			\hline
			\multirow{2}{1.7cm}{Bird's nest of tower}  & HSV-GLCM \cite{Xu2017HSV_GLCM_tower}  & PED & HSV-GLCM & Fusion & Test: 50 & Accuracy rate: 87.5\% \\
			\cline{2-7}
			& CF-CC \cite{Lu2018CF-CC_tower}  & ---- & CF-CC & Fusion & Train: 2972, test: 200 & Accuracy rate:  97.33\% \\
			\hline	
			\multirow{4}{1.7cm}{Broken strand of conductor} & CED-IFR \cite{Liu2012IFR_line} & CED & IFR & Shape & Test: 100 (10 fault images) & Recognition rate: 100\% \\	
			\cline{2-7}
			& LED-HT \cite{Yin2016LED-HT_line}  & LED-HT & Rules & Shape   & ---- & ----\\
			\cline{2-7}
			& CT \cite{Wang2015CT} & Gestal & Rules & Shape  & ---- & ---- \\
			\cline{2-7}
			& GVN-SWT \cite{zhang2019GVN_line} & GVN & SWT & Texture & Test: 400 & Accuracy rate: 85.5\% \\
			\hline			
			\multirow{3}{1.7cm}{Foreign body of conductor}  & DAG-SVM \cite{mao2019HOG_SVM_line}  & ---- & DAG-SVM & Shape & Train: 301, test: 34 & Accuracy rate: 84.3\% \\
			\cline{2-7}
			& SSD \cite{Wang2018SSD_line}   & ---- & SSD & Deep & Train: 4500, test: 1500 & Mean average precision: 85.2\% \\
			\hline	
			\multirow{2}{1.7cm}{Vegetation encroachment}  & PCNN \cite{Mills2010PCNN_line}  & ---- & PCNN & ---- & Test: 10 & Detection rate: 96\% \\
			\cline{2-7}
			& CNN-SM \cite{Qayyum2018Deep_stereo_line}   & ---- & CNN-SM & Deep & Test: 40 instances & Accuracy rate: 90\% \\
			\hline			
			\multirow{2}{1.7cm}{Broken of fitting}  & CED+HT \cite{Song2015CED-HT_fitting}  & CED+HT & Rules & Shape  & ---- & ---- \\
			\cline{2-7}
			& Faster R-CNN \cite{Tang2018Fst_fitting}   & ---- & Faster R-CNN & Deep & Train: 1000, test: 500 & Recall: 83.4\% \\
			\hline	
			\multirow{3}{1.7cm}{Missing pin of fitting}  & HM-LA \cite{Fu2017Haar_fitting}  & Haar+Adaboost & HT+LSD & Shape  & ---- & ---- \\
			\cline{2-7}
			& CNN \cite{Wang2018CNN_fitting}  & ACF+Adaboost & CNN & Deep & Train: 1900, test: 752 & Accuracy rate: 96.54\% \\
			\hline			
		\end{tabular}  
	}
\end{table*}

\subsubsection{Surface fault of Insulator}	
Some studies concentrated on the single surface fault of insulators such as icing and flashover.
Yang et al. \cite{yang2017IULBP} presented a classification method of ice types on insulators based on the texture feature descriptor.
According to the severity, they categorized the ice types into free of ice, glaze ice, heavy rime, medium rime, slight rime, partial rime and snow.
An improved uniform LBP (IULBP) was proposed for feature extraction.
Then, the extracted feature were compared with the predetermined template of the ice type.
Thus, the ice type of the insulator can be classified according to the similarity between the extracted feature and the predetermined template.
The authors evaluated their method at few images that were cropped to focus on the icing part.
Therefore, they excluded the insulator detection stage from general fault diagnosis framework.
Hao et al. \cite{Hao2018GSS-GSO_ins_ice} assessed the icing condition of insulators based on the geometric structure of the icing insulator.
The GrabCut was employed to segment the insulator from images.
The hand-craft rules were designed to classify the icing condition based on the distance properties between two neighbouring insulator caps.
These distance properties were defined as Graphical Shed Spacing (GSS) and Graphical Shed Overhang (GSO).
The method was tested by using 8 images and results showed it can recognize icing conditions quantitatively.
Zhai et al. \cite{Zhai2018M-SA_ins_flash} applied Faster Pixel-wise Image Saliency Aggregating (F-PISA) to detect insulators.
The flashover area in the detected insulator can be extracted based on the color determination in Lab color space.
The method was evaluated by using 100 insulator images with flashover fault and achieved 92.7\% detection rate.

Few researchers introduced the fault diagnosis scheme to determine multiple surface faults of insulators and most of them followed the same basic diagnosis procedure: detected the insulator first, then divided the insulator region into several parts, finally calculated the similarity between each part.
Oberweger et al. \cite{oberweger2014CGL-EGL_ins} extracted Difference of Gaussian key-points and calculated Circular GLOH-like (CGL) descriptor at each key-point.
The descriptors were reduced through Principal Components Analysis (PCA) and then classified by using RANSAC-based clustering approach for identifying the insulator.
Since the insulator region was detected, each caps can be separated from the insulator region by means of Grabcut segmentation and Canny edge detection.
Then, the Elliptical GLOH-like (EGL) descriptor was computed at every individual caps.
Finally, the faulty cap can be determined according to the Local Outlier Factor (LOF) between each cap.
The method was tested by using 400 aerial images with 20 faulty caps including 16 cracked caps and 4 flashover caps.
The true positive rate with 95\% that was outperformed the GLCM-based method which was introduced in \cite{Zhang2010GLCM_ins_fault}.
Zhao et al. \cite{zhao2016M-PDF_ins} presented a deep-learning-based method for the classification of the insulator status including normal, damaged, dust contamination and missing caps.
The insulator was detected by utilizing OAD-BSPK which was proposed in \cite{zhao2015OAD-BSPK}.
After insulator detection, the insulator region was divided into several parts.
Then, these sub-images as multiple image patches were resized to 256$\times$256 and can be input to the pre-trained AlexNet (a CNN framework for classification) for feature extracting.
Finally, the feature vector obtained from AlexNet with 4096-dimension can be classified by means of a trained SVM.
Experiments were conducted on 1000 samples with 98.71\% mAP.

\subsubsection{Missing-cap of Insulator}
The diagnosis of insulator missing-cap is a popular research issue in the power lines inspection domain.
The number of relevant literature is also the largest compared with other inspection tasks.
The main reason for this phenomenon can be attributed to the following points:
1) the insulator is widely used in the power lines and has significant function for mechanical support and electrical insulation.
2) the missing-cap of insulator occurs frequently.
3) the characteristic of missing-cap fault in the aerial image is invariable and obvious.

The missing-cap can be detect through out the partition based procedure that separates the insulator region into several parts and calculate their similarities.
Wang et al \cite{Wang2014ins_drop,wang2016CGT_LBP_HSV_ins} located the insulator by using the fusion feature based method which is introduced in the Section IV.A.
Then, the insulator region was rotated to horizontal and divided into 23 blocks.
The texture feature of GLCM was extracted from each block and used for similarity calculation.
Finally, the anomalous block was identified as the missing-cap region.
A sequential images of the diagnosis process demonstrated the performance of the proposed method.

The partition based procedure was limited by the size setting of the part and the repeated computation of the similarity calculation.
Therefore, some attempts have been made for missing-cap detection by using morphological operation in the whole insulator region to high light the faulty area.
Zhai et al. \cite{zhai2017S-AM_ins_drop} detected the missing-cap of insulators based on saliency and adaptive morphology (S-AM).
The insulator region was located by using saliency detection that combined with color feature and gradient feature.
Color model was used to segment the insulator from the located region for fault analysis.
The missing cap fault can be high lighted after the operation of adaptive morphology.
In experiments, the proposed method achieved 92.4\% detection rate on 100 aerial images and was compared with other competitive approaches (\cite{Wang2014ins_drop} with 65.4\%, \cite{Zhang2014ins_drop} with 85.7\%).
However, the S-AM can only deal with the fault of glass insulators.
To this end, authors improved the S-AM method in the study of \cite{zhai2018SMF_ins} to handle both glass and ceramic insulators.
They located the insulator by using color model and rotated the insulator into horizontal.
Then, the morphological operation was performed to obtain the projected curve of fault features.
Finally, according to the hand-craft rules, the fault position can be determined.
Experiment results demonstrated the ability of the proposed method (92.8\%) compared with S-AM (92.4\%) \cite{zhai2017S-AM_ins_drop}.
Han et al \cite{Han2019MYOLO_AM_ins} also diagnosed the missing-cap by utilizing morphological operation. 
The modified YOLOv2 detection framework was introduced to detect insulators.
Similar to the research in \cite{zhai2017S-AM_ins_drop}, they used adaptive morphology to high light the fault region of missing-cap.
But in the segmentation of the insulator, the color model combined with GrabCut was applied rather than the color model.
Totally 120 images (42 original images augment to 120 processed images) with missing-cap of insulators were used to test the proposed method.
In competitively experiments, the researchers compared their method with S-AM \cite{zhai2017S-AM_ins_drop} and SMF \cite{zhai2018SMF_ins} that mentioned above and achieved the best performance with 96.3\% precision and 93.4\% recall.

Recently, deep learning had attracted considerable interests in the power lines inspection and most of the studies concentrated on the detection of insulator and its fault
\cite{tao2018ILN_DDN_ins},
\cite{ling2018Fst-Unet},
\cite{li2018insulator_rfcn},
\cite{sampedro2019UpNet_CNN_ins},
\cite{liu2018frcnn_ins, yang2019DCNN_ins, gao2017Fst_FCN_for_ins,  tian2018Parallel_ins, jiang2019EL-MLP, chen2019SOFCN}.
For example, Ling et al. \cite{ling2018Fst-Unet} applied Faster R-CNN to detect the insulator and employed U-Net to segment the missing-cap fault area in the detected region.
The method was evaluated by using 55 faulty images and achieved 95.1\% precision and 95.5\% recall.
Li et al. \cite{li2018insulator_rfcn} detected missing-cap by using Region based Fully Convolutional Network (R-FCN).
The training and testing sets composed of 2626 and 500 respectively and the method achieved 90.5\% AP.
Sampedro et al. \cite{sampedro2019UpNet_CNN_ins} proposed a Up-Net to segment the insulator and constructed a 10-layer CNN to determine the missing-cap fault.
For training and testing the diagnosis model, 2400 and 400 images were used, and the method obtained the accuracy of  98.8\%.
More details about these deep learning based approaches will be further discussed in section V.B.

\subsubsection{Corrosion of tower}	
A few examples can be found on the corrosion determination of power towers by using the closely photographed image.
Maeda et al. \cite{maeda2017DELM-LRF} estimated the corrosion level of the transmission tower based on Local Receptive Field (LRF) and Deep Extreme Learning Machine (DELM).
The research focused on surface images of the tower and applied LRF to extract features for further diagnosis.
The LRF functioned as CNN that performed convolution and pooling in the input image.
Then, the DELM was utilized to classify the extracted features into three corrosion levels.
Totally 2797 images with 5-fold cross validation were utilized in the experiment and 79.6\% F-measure was achieved.
In the research \cite{maeda2018CMDELM-LRF}, the authors modified the DELM-LRF \cite{maeda2017DELM-LRF} by combining the text feature.
Text information such as type of towers, height of towers, voltage level and coating year was translated into a feature vector, and then was inputted to the framework of DELM-LRF with visual feature simultaneously. 
In the experiment, totally 3017 samples with 5-fold cross validation was utilized.
The performance with 88.8\% F-measure of the modified DELM-LRF (defined as Correlation-Maximizing DELM-LRF) showed a great improvement compared to DELM-LRF.

\subsubsection{Bird's nest of tower}	
Studies presented in the following discussed the detection of bird's nest on the power tower which is similar to the common object detection task.
Xu et al. \cite{Xu2017HSV_GLCM_tower} presented a bird's nest detection method for transmission towers.
In the detection stage, the tower region was located by using Prewitt direction operator and hand-craft rules.
For fault diagnosis, the image region of the tower was converted to HSV color space and the candidate regions were identified based on the color model.
GLCM was calculated at each candidate region to analyze the texture feature of bird's nest and then the fault can be detected.
Experiments were conducted on 50 aerial images and 87.5\% detection rate was achieved.
Contrast to \cite{Xu2017HSV_GLCM_tower}, the research in \cite{Lu2018CF-CC_tower} removed the detection stage and directly located the bird's nest.
The nest suspected region can be identified by using local adaptive binarization and template convolution.
Then, a cascade classifier was established to determine the correct nest region.
This cascade classifier was constructed by 3 SVMs including: SVM-1 with trunk feature, SVM-2 with projection features and SVM-3 with improved burr feature.
In the comparison experiments, 2972 and 200 images were used for training and evaluation. 
Results indicated the obvious improvement of the proposed method (97.33\% accuracy) compared with HSV-GLCM \cite{Xu2017HSV_GLCM_tower} (61.85\%) that mentioned above.

\subsubsection{Broken strand of conductor}	

A few attempts have been made to detect the broken strand of conductor and most of them follow the similar framework: extract the line segment and then determined the abnormal segment by using hand-craft rules.
Liu et al. \cite{Liu2012IFR_line} dealt with broken strand of the transmission conductor based on Improved Freeman Rule (IFR).
Canny Edge Detector was applied to extract segments in the input image.
According to the extracted segments and the characteristics of the end point, the conductor can be rotated to horizontal.
Then, the IFR was used to determine whether there exists broken strand in the detected conductor.
The proposed method was tested by using 100 images and all broken strand faults were correctly recognized.
Yin et al. \cite{Yin2016LED-HT_line} applied Laplacian Edge Detector (LED) combined with Hough Transformation (HT) to extract the lines.
Based on the extracted lines, the region of conductors can be located by employing the Region Growing.
As for the fault identification, the hand-craft rules were designed on the basis of the width change of the detected conductor.
Results on a few images were illustrated the performance of the proposed method.
Wang et al. \cite{Wang2015CT} employed Cross Template to detect vertical and horizontal lines.
The extracted lines were grouped based on the Gestalt perception theory.
According to the different perceptual contours, the fittings such as dampers and spacers installed at the conductor can be filtered out.
Finally, similar to the research in \cite{Yin2016LED-HT_line}, the hand-craft rules were established to recognize the broken strand.
Besides the width change, the proposed rules contained more parameters such as absolute gray difference and relative gray difference.
The presented method was evaluated at several images and the performance was demonstrated by some visualized results. 

Different to the aforementioned studies, Zhang et al. \cite{zhang2019GVN_line} established a monitoring system of transmission conductor based the texture structure on the conductor surface.
In the image analysis algorithm of the monitoring system, the aerial image should be converted to gray color space by using Gray-scale Variance Normalization (GVN).
Then, the conductor can be extracted based on adaptive threshold segmentation with morphological processing.
The gray value distribution of the conductor region can be represented based on the Square Wave Transformation (SWT).
According to the characteristic of the conductor, the broken strand would break the repeated helical structure of the normal conductor.
Thus, the broken strand can be identified by analyzing the Z-shaped waveform from SWT.
The proposed method achieved 90.5\% accuracy in 400 aerial images with simple background, and 85.5\% in 400 images with complex background.

\subsubsection{Foreign body of conductor}

The procedure of the foreign body detection was similar to the inspection task of bird's nest detection.
Mao et al. \cite{mao2019HOG_SVM_line} detected the foreign body of the conductor based on HOG and SVM.
Firstly, the aerial image was processed by gray-scale and median filter for further analysis.
Then, the HOG feature was extracted and classified by Directed Acyclic Graph (DAG) multi-classifiers that defined as DAG-SVM.
The DAG-SVM consisted of three SVM classifiers that responsible for different categories.
For classifier 1, the unusual image and Non-foreign-body image were distinguished and inputted to next two classifiers respectively.
The classifier 2 was utilized to determine whether the unusual image belonged to foreign body or broken strand.
The rest classifier recognize the Non-foreign-body image into two categories: broken strand and normal.
Finally, the condition of the transmission conductor can be obtained.
In experiments, 335 images were utilized with 10-fold cross validation for training and testing.
The recognition accuracy with 84.3\% illustrated the effectiveness of the proposed method.
Tang et al. \cite{Wang2018SSD_line} presented a deep-learning based method for foreign body detection.
The object detection framework SSD that employed the VGG (a CNN for classification) as the basic network was applied to detect kite, balloon and bird's nest in the power lines.
Each type of foreign body had 1500 training samples and 500 testing samples with 300$\times$300 resolution.
Authors discussed the parameter setting of the detection method, results showed the box ratio with \{1/2,2\} and training batch size with 4 can achieve better performance with 85.2\% mAP.
The competitive experiments also conducted with the shallow-feature-based detection framework such as DPM that achieved 54.8\% mAP.
It demonstrated the powerful capabilities of the proposed deep-learning-based method.

\subsubsection{Vegetation encroachment of conductor}

The visible image analysis of vegetation encroachment is quite different to other inspection items, it should be combined with distance measurement instead of object detection or classification alone.
The commonly used approach for distance measurement in the optical based aerial inspection was binocular stereo vision.
To determine the vegetation encroachment, the vegetation (trees) and transmission conductors should be located manually or automatically first, and then the distance between them can be estimated.
Mills et al. \cite{Mills2010PCNN_line} segmented the crown of trees in the multi-spectral image by using Pulse-Coupled Neural Network (PCNN) and morphological operation.
The horizontal distance between the conductor and trees along with the height of trees and towers can be estimated by stereo vision.
The stereo image was obtained from subsequent frames of a single camera that had the same effectiveness with the binocular camera.
To obtain depth information in the stereo image, a stereo matching algorithm was proposed based on the dynamic programming.
In the experiment, the detection rate of tress reached 96\% in 10 images with totally 129 trees.
The average error in the estimation of tree-line distance achieved 0.7 m.
And the height estimation of trees and tower attained 1.8 m and 1.1 m average error respectively.
Qayyum et al. \cite{Qayyum2018Deep_stereo_line} also applied the stereo image for monitoring vulnerable zones near transmission conductors.
However, the automatic detection of the trees is not the objective of this research, the authors paid more attention to the height estimation based on stereo vision.
For obtaining the stereo image, the binocular camera was installed on a fixed wing UAV.
In order to calculate the height of objects proximal to transmission conductors, they presented a 8-layer CNN for Stereo Matching (CNN-SM).
The experiment was implemented in a 500 kV power corridor which comprised 20 towers.
The proposed method was compared with existing algorithms such as dynamic programming and graph cut and achieved higher accuracy of 90\%.

\subsubsection{Broken of fitting}	
A few examples can be found on the detection of broken fittings.
Song et al. \cite{Song2015CED-HT_fitting} applied Canny edge detector combined with Hough transform to extract the edge of the conductor.
Next, along the direction of the conductor, a scanning window was established.
Then, the candidate region of spacers can be recognized by finding the minimum white area in all the windows that slid through the conductor.
Finally, the hand-craft rules were designed based on connected components calculation to identify whether the detected spacer was broken.  
If the number of connected components is lager than 1, the spacer was recognized as broken spacer.
As results, a sequence of the visualized images in the algorithm procedure illustrated the effectiveness of the proposed method.
Contrast to \cite{Song2015CED-HT_fitting}, Tang et al. \cite{Tang2018Fst_fitting} treated the broken fitting detection as a conventional detection task.
They employed Faster R-CNN to detect broken dampers and other normal fittings.
Inspection images with 5 categories were prepared for training and validation including: two types of the spacer, normal damper, broken damper and bird's nest.
There were 1000 training samples and 500 testing samples for each category in the experiments.
Authors discussed the performance of the proposed method under different situation.
Result of 83.4\% recall demonstrated the basic network with ResNet and convolutional kernel size with 9$\times$9 performed better.

\subsubsection{Missing pin of fitting}	
The challenging inspection task of missing pin diagnosis has only been investigated in few studies due to the extremely small size.
The detection of the small fitting such as pin and nut is still an opening issue, thus, these studies analyzed the missing pin based on the aerial image that was captured close to the fitting or even cropped the fitting region from original image manually.
Fu et al. \cite{Fu2017Haar_fitting} introduced a hierarchical model with learning algorithm to identify the missing pin.
According to the And-or Graph (AoG), the fitting can be represented by the combination of several parts.
For example, the fastener can be divided into two parts: pin and nut.
In order to detect each part of the fitting, the Haar-like feature and Adaboost classifier were applied.
For missing pin identification, the detected fitting region was processed with LSD and Hough transform to extract segments and circles respectively.
Then, the missing pin fault can be identified based on the distance constraint between the center of the circle and the segment of the pin.
This method was tested by using 42 images of fitting region, and 5 images were considered have pins while only one of them was correct.
Wang et al. \cite{Wang2018CNN_fitting} proposed a CNN based method for missing pin diagnosis.
The fitting region was located by using Aggregate Channel Features and Adaboost classifier.
Then, a 8-layer convolutional neural network was established to extract deep features of the fitting region and classify them into three categories: normal fitting, fitting with missing pin and background.
The diagnosis method was trained by 1900 images and evaluated at 752 images and achieved 96.54\% recall.
However, the faulty image was already the fitting region cropped by hand, which meant the jointly experiment was not conducted in this research that performed detection and diagnosis in-order.

\subsubsection{Remarks}

\mbox{Table. \ref{tab:summary_diag}} provides the valuable information of researches in power line fault diagnosis, which includes the fault category, proposed method, approach used in component detection stage, approach used in fault identification stage, main image features, brief description of data, and the method performance.

The fault diagnosis of power line is a relatively rare touched area in the literature compared with component detection.  
The problems in this area is similar to the component detection to some extent, but there are
still several nuances should be concerned.
Current researches mainly treat the fault diagnosis as object detection task (e.g., missing-cap of insulator) or classification task (e.g., corrosion of tower).
In reality, one fault has various forms that leads to the difficulties in robust algorithm design. 
It is worth trying to identify the fault from the perspective of abnormal image detection .
There is a primary attempt in \cite{sampedro2019UpNet_CNN_ins} to classify abnormal images.
For fault types, the missing-cap of insulator received most of the attention while the works of other faults are limited.
In addition, we find that in most cases, one paper only focused on one fault of a specific component.
With the widely application of aerial inspection and the accumulation of inspection data, more fault types need to be considered.
As for image features, shape and deep features are most frequently used in the existing literature.
Since the fault data is relatively rare, the hand-craft extractor for fusion features would need some further attention.
Moreover, using multi-modal learning to leverage the rich information of text data is another good choice which is preliminary tried in \cite{maeda2018CMDELM-LRF}.
To identify the fault, most studies need to detect the component region first.
However, few researchers take it into consideration that how to achieve fault identification when the component is miss detected.
Fault diagnosis without the stage of component detection deserves further investigation.
In addition, we also find that most existing works are evaluated in laboratory.
More real-world experimental results in practical aerial inspection of power lines are welcomed for the research.

\subsection{Main limitations of current researches}
Although the power lines inspection has developed rapidly in recent years, there are still two main limitations in the existing literature that need some further attention.
The first is the insufficient research on some power line components and their faults.
As can be seen from the previous review, most of the research is focused on the insulator together with its faults while other components are only received rare attention.
The reasons for this phenomenon are as follow:
In four categories of crucial components, the insulator has lowest variants in images due to its standardized shape, which makes the algorithm easy to get higher generalization in real-world applications.
Further, the insulator has moderate size in aerial images while the tower is too large, the conductor is overly thin and the fitting is excessively small.
This factor results in conveniently photographing for UAVs to capture more images of insulators.
In addition, the moderate size and standardized shape also reduces the difficulty in method design.
Finally, components apart from insulators have many variants or subcategories or scales in aerial images.
For instance, the damper, fastener and spacer all belongs to fittings and they are very different in size and shape.
The variants, insufficient data and inappropriate scale make researches on other components is a rarely touched area in the literature. 

The second is that most methods in current works have not been tested in actual engineering.
In laboratory, the data collected from aerial inspection is separated into training set and testing set, which means they are identically distributed.
But in reality, this precondition can not be guaranteed.
Generally, the appearances of component, fault and background have a lot of variants in real world inspection image, and some variants are not included in the experimental data. 
Moreover, the image differences between the lines in different regions are even greater.
This challenging problem places higher requirements on the robustness and generalization capability of both hand-craft designed methods and learning based methods.
Nevertheless, the researches on effective evaluation of the robustness and generalization of the analysis methods are still limited.
Another factor which limits the practical application of inspection data analysis is the computation cost.
There are massive images and videos with high pixel resolution need to be analyzed within an inspection period.
Under the situation of limited computing resources, the analysis method should achieve highly efficient computation.
However, researches on acceleration of the analysis model for inspection data are quite rare. 
In addition, the computation time of the analysis method is rarely introduced in the existing literature.



\section{Deep-learning-related analytic methods in power lines inspection}
Deep learning has been widely used in generic tasks such as car detection and face recognition, and its application in power lines inspection is becoming a research hotspot in the past two years.
In this section, with the objective to offer an in-depth discussion of current deep-learning-related researches in power lines inspection, we summarize these works (some of them are briefly mentioned above) with special attention paid to their method characteristics, research issues and core ideas.
Firstly, we provide a brief introduction of some fundamental deep learning approaches for batter understanding the deep-learning-based researches in the field of power lines inspection.
Then, the exploration and taxonomy of current deep-learning-related methods for the inspection of power lines are introduced from five aspects: using existing frameworks, extracting deep features, network cascading, aiming at data insufficiency, and improving methods based on domain knowledge.
Valuable information of the literature is listed in \mbox{Table. \ref{tab:summary_DL}}.
Finally, we propose a basic conception about how to conduct an intelligent analysis system of inspection data by using the deep learning technology and several novel image processing approaches, some alternative methods are also provided in each stage of the system.


\subsection{A brief introduction to fundamental deep learning approaches}

\subsubsection{Deep convolutional neural network}
Deep convolutional neural network (DCNN) has the capability to extract high quality features and is widely used in a variety of tasks.
It has made great achievements in the field of computer vision (e.g., image classification), and outperforms other Non-DCNN based algorithms.
A typical DCNN consists of multiple layers which aims to learn the representation of input data.
Most layers of DCNN are composed of a number of feature map, within which each unit acts like a neuron.
There are three major types of layers in DCNN: convolutional layer, pooling layer and fully connected layer.
In the convolutional layer,  units of the feature map are connected to local patches in the feature maps of the previous layer through the 2D convolutional kernel (or filter or weights).
The role of the pooling layer is to downsampling of feature maps.
The fully connected layer provides the feature vector for classifiers (e.g., SVM or Softmax).
A typical CNN is composed of several stacked convoluitonal and pooling layers, followed by the fully connected layer.

Since the appearance of AlexNet \cite{krizhevsky2012alexnet}, a lot of novel DCNN architectures have been proposed by restructuring the processing unit and designing the new block.
ZF-Net \cite{zeiler2014ZFNet} and VGG-Net \cite{simonyan2014vgg} increased the depth of the DCNN by reducing the size of the filters.
GoogleNet \cite{szegedy2015googlenet} reduced the computational cost through inception block.
In 2015, the residual block (or skip connections) was proposed in ResNet \cite{He2015ResNet} which got famous.
This concept of skip connections was utilized by many succeeding DCNN architectures such as Inception-ResNet \cite{szegedy2017inc-res},and ResNext \cite{xie2017ResNext}.
Some researchers concentrated on the lightweight DCNN for mobile device such as MobileNet \cite{howard2017mobilenet}, Xception \cite{chollet2017xception}, and SuffleNet \cite{zhang2018shufflenet}.
Recently, some attempts have been made to automatically design the DCNN architecture (also known as Neural Architecture Search) such as NasNet \cite{zoph2018nasnet}, MNasNet \cite{tan2019mnasnet}, and ENas \cite{pham2018enas}.

\subsubsection{Dee Learning Based Object Detection and Segmentation}
The object detection method based on deep learning consists of two parts: a DCNN (also defined as backbone or basic network) for feature extraction and a detecting scheme for object classification and location.
According to the detecting scheme, the DL-based detection method can be summarized into two major categories \cite{liu2018deep_OD_survey}:
(1) Two-stage detection method which needs to generate proposals of possible objects in an independent stage.
The proposal can be regarded as a specific bonding box that may have a object and be generated from an image.
In a two-stage detection method, deep features are extracted from these proposals, and then classified by category-specific classifiers.
The classic and probably the most commonly used method is Faster R-CNN, introduced by Ren et al. \cite{ren2015faster-rcnn} in 2015.
Many remarkable methods have emerged in the same period such as R-FCN \cite{dai2016rfcn}, Cascade R-CNN \cite{cai2018cascade_rcnn}, and Light Head R-CNN \cite{li2017Light_head_r-cnn}.
(2) One-stage detection method which does not contain the generation of proposals.
For adjusting to the mobile device that has limited storage and computational capability, the one-stage detection method removes the procedure of proposal generation and its subsequent feature processing operations (e.g.,  classification).
As an alternative, the method directly obtains the category and position information from preset grids of the full image with a single DCNN.
Commonly used methods are SSD \cite{liu2016ssd}, YOLO\cite{redmon2016YOLO,redmon2018yolov3}, and RetinaNet \cite{lin2017retinanet}.


Recently, some studies opened up a new direction of DL-based object detection method which is called anchor-free detection method.
These methods utilized a key-point-like approach to represent the position of objects instead of a traditional bounding box or anchor.
The popular anchor-free methods are CornerNet \cite{law2018cornernet}, ExtremeNet \cite{zhou2019ExtremeNet}, CenterNet \cite{duan2019centernet}, and FCOS \cite{tian2019fcos}. 

In addition to aforementioned object detection methods, the segmentation method also has the function to detect objects in an image.
In segmentation methods, each pixel is classified with the category of its enclosing object.
There are some commonly used methods that have become widely known standards such as FCN \cite{long2015FCN}, U-Net \cite{ronneberger2015UNet}, and SegNet \cite{badrinarayanan2017segnet}.
Recently, some attempts (e.g., Mask R-CNN \cite{he2017mrcnn} and HTC \cite{chen2019HTC}) have been made to combine the object detection and segmentation that is called instance segmentation.
These methods achieved the label separation for different instances of the same category. 
In other words, they can achieve pixel-wise classification in each bounding box that contains the object.

\subsubsection{Generative Adversarial Networks}
Generative Adversarial Networks (GANs) have attracted widespread attention especially in computer vision field which was proposed by Goodfellow et al. \cite{goodfellow2014GAN} in 2014.
GANs consist of two networks and train them in competition with each other, these two networks are described as follow:
a network so-called generator is utilized to generate synthetic data samples, 
another network so-called discriminator is used to distinguish real data samples from synthesized samples.
Due to the capacity of new data generation from the learned statistical distribution of training data, GANs achieved state-of-art performance in various vision applications including image synthesis, segmentation, style transfer, and image super-resolution.

Since original GANs, there are many variants in different fields have been proposed.
Some studies focus on generating high-quality samples such as CGAN\cite{mirza2014CGAN}, DCGAN \cite{radford2015DCGAN}, and WGAN\cite{arjovsky2017WGAN}.
Few attempts have been made to image style transfer, i.e. converting images from one style to another such as day to night.
The typical researches include Pix2Pix \cite{isola2017Pix2Pix}, and CycleGAN \cite{zhu2017cycleGAN}.
GANs are also widely used in image restoration and the well-know researches are DeblurGAN \cite{kupyn2018DeblurGAN} and SRGAN \cite{ledig2017SRGAN}.




\subsection{An exploration of current deep-Learning-based approaches for the inspection of power line components}
\begin{table*}[ht]  
	\caption{Summary of the related work of deep-learning-based approaches for the inspection of power line components.}  
	\label{tab:summary_DL}
	\centering
	\renewcommand\arraystretch{1.2}
	\setlength{\tabcolsep}{1.5mm}{
		\begin{tabular}{m{1.8cm} l l c c m{5cm}} 
			\hline
			\textbf{Characteristic} & \textbf{~~~Inspection item}  & \textbf{~~~~~~Method} & \textbf{Data size}& \textbf{Pixel size} & \textbf{~~~~~~~~~~~~~~~~~~Core idea} \\
			\hline
			\multirow{9}{1.8cm}{Existing frameworks} & Missing-cap of insulator & Faster R-CNN \cite{liu2018frcnn_ins} & 4500 & 500$\times$500 & Utilize Faster R-CNN to detect insulator and it's fault \\
			\cline{2-6}
			& Tower detection & Faster R-CNN \cite{bian2019Fst_tower} & 1300 & 640$\times$480 & Utilize Faster R-CNN to detect tower \\
			\cline{2-6}
			& Conductor detection & FCNs \cite{hui2018Fst_FCNs} & 600 & 1280$\times$720 & Utilize FCNs to detect power line \\
			\cline{2-6}
			& Fitting detection & Faster R-CNN \cite{wang2017Fst_for_fitting} & 6000 & 500$\times$500 & Utilize Faster R-CNN to detect fittings\\
			\cline{2-6}
			& Insulator detection & YOLO \cite{wang2018YOLO_for_ins} & 1000 & 448$\times$448 & Utilize YOLO to detect insulator \\
			\cline{2-6}
			& Tower detection & YOLOv3 \cite{chen2019YOLO_tower} & 13429 & 352$\times$352 & Utilize YOLO to detect tower \\
			\cline{2-6}
			& Insulator detection & SSD \cite{xu2018SSD_for_ins} & 2500 & 512$\times$512 & Utilize SSD to detect insulator \\
			\cline{2-6}
			& Conductor detection & cGAN \cite{chang2018cGAN_for_line} & 5500 & 256$\times$256 & Utilize cGAN to detect conductor \\
			\cline{2-6}
			& Insulator detection & cGAN \cite{chang2018cGAN_for_ins} & 3000 & 256$\times$256 & Utilize cGAN to detect insulator \\
			\hline
			\multirow{5}{1.8cm}{Extracting deep features} & Surface-fault of insulator & M-PDF \cite{zhao2016M-PDF_ins} & 1000 & 227$\times$227 & Extract features by CNN in multi image patches\\
			\cline{2-6}
			& Corrosion of tower  & CMDELM-LRF \cite{maeda2018CMDELM-LRF} & 3017 & 50$\times$50 & Extract features by CNN in image and text\\
			\cline{2-6}
			& Missing-cap of insulator  & DCNN \cite{yang2019DCNN_ins} & 2951 & 256$\times$256 & Extract features by CNN in sub-windows of aerial image\\
			\hline
			\multirow{5}{1.8cm}{Network cascading} & Missing-cap of insulator & \tabincell{l}{Faster R-CNN \\+ U-net\cite{ling2018Fst-Unet}} & 620 & 1024$\times$1024 & Utilize Faster R-CNN to detect insulator and U-net to detect the fault \\
			\cline{2-6}
			& Missing-cap of insulator  & \tabincell{l}{Faster R-CNN \\ + FCN \cite{gao2017Fst_FCN_for_ins}}  & 3650 & 1215$\times$1048 & Utilize Faster R-CNN to detect insulator and FCN to filter out background \\
			\cline{2-6}
			& Missing-cap of insulator  & ILN + DDN \cite{tao2018ILN_DDN_ins} & 1956 & ---- & Propose an Insulator localizer network and a Defect detector network\\
			\hline	
			\multirow{7}{1.8cm}{Aiming at data insufficiency} & Insulator detection & \tabincell{l}{Synthetic method \\ + cGAN \cite{chang2018Synthetic_ins}} & 265 & 512$\times$512 & Propose a synthetic method to synthesize training samples \\
			\cline{2-6}
			& Missing-cap of insulator & PPM \cite{tian2018Parallel_ins} & ---- & ---- & Introduce a preprocessed parallel method by data augmentation\\
			\cline{2-6}
			& Surface fault of insulator & SPPNet-TL \cite{bai2018SPPNet-TL_ins}  & 278 & 227$\times$227 & Training on a small dataset based on transfer learning\\
			\cline{2-6}
			& Insulator detection & SSD + TS-FT \cite{miao2019TS-FT_ins}  & 8005 & 300$\times$300 & Introduce a two-stage fine-tune strategy for training on the small dataset   \\
			\cline{2-6}
			& Conductor detection & WSL-CNN \cite{lee2017weakly_line} & 8400 & 512$\times$512 & Apply weakly surprised learning to train the conductor detection model\\
			\hline	
			\multirow{5}{1.8cm}{Improving by domain knowledge} & Missing-cap of insulator  & EL-MLP \cite{jiang2019EL-MLP} & 485 & 300$\times$300 & Aggregate deep learning models in perception levels based on ensemble learning \\
			\cline{2-6}
			& Missing-cap of insulator & SO-FCN \cite{chen2019SOFCN} & 300 & 400$\times$600 & Introduce a mathematical morphology operation to optimize the detection procedure \\
			\cline{2-6}
			& Missing-cap of insulator & Up-Net + CNN \cite{sampedro2019UpNet_CNN_ins} & 2800 & 256$\times$256 & Propose a diagnosis strategy for missing-cap detection based on semantic segmentation\\
			\cline{2-6}
			& External force damage & \tabincell{l}{Modified\\Faster R-CNN \cite{xiang2018Modified_Fst_ins}} & 2199 & 600$\times$1000 & Improve Faster R-CNN to detect engineering vehicles based on their characteristics \\
			\hline 
		\end{tabular}  
	}
\end{table*}

\subsubsection{Directly use of existing frameworks}

Faster R-CNN is a common used framework in the inspection of power lines for insulator fault detection \cite{liu2018frcnn_ins}, tower detection \cite{bian2019Fst_tower,hui2018Fst_FCNs}, and fitting detection \cite{wang2017Fst_for_fitting}.
Liu et al. \cite{liu2018frcnn_ins} applied Faster R-CNN to detect the insulator and the missing cap fault separately. 
They tested the method with insulator images in three different voltage level and prepared 1000 training samples and 500 testing samples for each level.
For the diagnosis of missing cap fault, only 120 images (80 for training) were utilized for evaluation.
In the experiment, the all the images were resized to 500$\times$500 pixel resolution and data augmentation including flipping and cropping was applied to extend the dataset.
Bian et al. \cite{bian2019Fst_tower} used Faster R-CNN for tower detection.
Totally 1300 aerial images were prepared for experiments and the 10-fold cross-validation was applied to find best model.
Hui et al. \cite{hui2018Fst_FCNs} also employed Faster R-CNN to locate towers.
Furthermore, the conductor was extracted by using FCNs.
The data with 1280 tower images (1000 for training) and 600 conductor images (400 for training) was used in experiments.
Wang et al. \cite{wang2017Fst_for_fitting} applied Faster R-CNN to detect fittings including space, damper and arcing ring.
For each type of fittings, 1500 training samples and 500 testing samples were prepared and all images were resized to 500$\times$500 pixel resolution.

In order to achieve high computation speed, the one stage detection framework was applied in some researches such as YOLO \cite{wang2018YOLO_for_ins,chen2019YOLO_tower} and SSD \cite{xu2018SSD_for_ins}.
Wang et al. \cite{wang2018YOLO_for_ins} employed YOLO to detect insulator in the image with gray color space.
The data including 1000 images (800 for training) was collected in laboratory and outdoor power lines.
All the images were resized to 448$\times$448 for matching the input size of the network.
Chen et al. \cite{chen2019YOLO_tower} utilized the improved YOLO (also denoted as YOLOv3) to detect towers.
On account of the data insufficiency, they constructed a dataset by generating the simulated images.
Among 13429 images were used in the experiment, of which 11,951 for training and 1478 for testing, the pixel resolution for the network input was 352$\times$352.
The authors discussed that the pixel size of the input image was an important factor to influence the method performance.
Xu et al. \cite{xu2018SSD_for_ins} proposed a SSD based method for insulator detection.
Totally 2000 images were augmented by rotation and extended to the number of 6000 (500 for validation).
In experiments, the pixel resolution with 512$\times$512 showed higher accuracy compare to 300$\times$300 while both of them achieved the requirement of real-time detection.

Few studies applied unconventional detection framework (cGAN) to detect the power line components \cite{chang2018cGAN_for_line,chang2018cGAN_for_ins}.
Chang et al. \cite{chang2018cGAN_for_line} recognized the power conductor by using cGAN.
The aerial image was inputted to cGAN and the mask image that only contained the conductor was generated.
The pixel resolutions of the input and the output were 256$\times$256 and 128$\times$128 respectively.
Three datasets were prepared in experiments including training set with 5000 images, simple testing set with 500 images and difficult testing set with 500 images.
The authors also employed the cGAN for insulator detection \cite{chang2018cGAN_for_ins}.
A two-stage training strategy was proposed to obtain a more accurate cGAN model.
In the training stage 1, the model was trained by using the position samples with coarse annotation.
Then, the same model was continue trained by utilizing the segmentation samples with fine annotation.
Among 3000 images collected from the Internet were used for evaluation.
The input and output of the cGAN had the same pixel resolution with 256$\times$256.

\subsubsection{Extract deep feature for classification or detection}

A few examples can be found on the use of extracting deep feature for classification \cite{zhao2016M-PDF_ins,maeda2018CMDELM-LRF} or detection \cite{yang2019DCNN_ins}.
Zhao et al. \cite{zhao2016M-PDF_ins} classified the condition of insulators by means of AlexNet.
The deep feature was extracted by the untrained AlexNet which was pre-trained in the ImageNet dataset.
Then, the extracted deep feature was fed to a SVM for final classification.
Totally 1000 images with 256$\times$256 pixel resolution were used for training (70\%) and testing the proposed method.
On the purpose of deterioration Levels estimation for towers, Maeda et al. \cite{maeda2018CMDELM-LRF} extracted visual features by using LRF which performed convolution and pooling similar to CNN.
Different from the traditional image-based research, they combined with the text feature that was extracted by a hidden layer.
Two kinds of features are further extracted and classified by DELM.
In the experiment, 3107 images with 50$\times$50 pixel resolution were used and 5-fold validation was applied as verification method.
Yang et al. \cite{yang2019DCNN_ins} established a 9-layer CNN to extract deep feature from sub-windows of the original image for insulator fault detection.
For example, an aerial image with 1280$\times$720 pixel resolution can be divided into 15 small images by the adaptive sliding window.
Then, the sub-window can be determined by the CNN into two classes: normal and abnormal.
The CNN model was trained by 2610 sub-windows that obtained from 205 raw images and tested in 341 real-world images.

\subsubsection{Network cascading for fault diagnosis}

Studies presented in the following have discussed the structure of network cascading and they have concentrated on the fault diagnosis.
The network cascading was generally composed of two sequential deep learning networks such as the combination of Faster-CNN and U-net \cite{ling2018Fst-Unet}, the combination of Faster-CNN and FCN \cite{gao2017Fst_FCN_for_ins}, and the combination of Insulator localizer network (ILN) and Defect detector network (DDN) \cite{tao2018ILN_DDN_ins}.

The procedure of the network cascading greatly narrowed the scope of fault analysis in which the former network was responsible for component detection and the latter identified the fault on the located component region.
Ling et al. \cite{ling2018Fst-Unet} detected the insulator by using Faster R-CNN.
Then, the insulator region was cropped from the original image and inputted to the U-net for locating the missing cap.
In the experiment, 620 aerial image of 1024$\times$1024 pixel resolution were utilized for Faster R-CNN with 3-fold validation.
For training and testing of U-net, 220 insulator images contained missing-cap faults were cropped from original images and the 4-fold validation was performed.
The pixel resolution of the cropped image was various depending on the size of the insulator.
Gao et al. \cite{gao2017Fst_FCN_for_ins} also applied Faster R-CNN to detect insulators.
But instead of detecting the fault by U-net, they employed FCN to segment the insulator from the detected region.
Then, each cap of the insulator can be recognized for further fault identification.
Among 3000 aerial images with 1215$\times$1048 pixel resolution were utilized to train Faster R-CNN and 100 images were used for evaluation.
Due to labeling cost, only 450 and 100 insulator images with 500$\times$500 pixel resolution were prepared for FCN training and testing respectively.
Tao et al. \cite{tao2018ILN_DDN_ins} proposed the ILN and DDN to detect insulators and their fault based on different backbone (VGG and ResNet respectively).
The ILN first detect all the insulators in the aerial image, and then the detected regions were cropped and fed into the DDN for locating the missing-cap fault.
For the experiment, totally 900 normal images and 60 faulty images were acquired from UAV.
Due to the data insufficiency, the image synthetic algorithm and the data augmentation process were applied.
The image synthetic algorithm employed U-net to segment the insulator and then pasted it into other images with various backgrounds.
The data augmentation contained 7 image processing operations such as rotation, shift, shear, and shear.
Eventually, among 1956 images (1186 for training) for ILN and 1056 images (792 for training) with missing-cap fault were prepared.

\subsubsection{Objective to solve data insufficiency}

Data insufficiency is a challenging problem in the data analysis of power line inspection.
Some attempts have been made in the previous articles such as image synthesis (e.g., \cite{tao2018ILN_DDN_ins}) and data augmentation (e.g., \cite{liu2018frcnn_ins,xu2018SSD_for_ins}).
The following researches made some further investigation about the problem of data insufficiency.
Chang et al. \cite{chang2018cGAN_for_ins} employed the cGAN for insulator detection.
Due to the difficulty to obtain the real-world aerial images, they proposed a synthetic method to generate synthetic images from 65 real-world insulator images. 
The insulator region was overlapped to various background images with different parameters such as gaussian noise and transparency.
The synthetic dataset included three sample categories: sample with insulators, sample without insulators and sample with pseudo targets.
The cGAN model was trained by using 8000 synthetic images and tested in 200 real-world insulator images.
Both the input and output of the model had the same pixel resolution with 512$\times$512.
Tian et al. \cite{tian2018Parallel_ins} proposed a parallel method to solve the insufficient diversity of acquired inspection data.
The original input image was processed with different operation (e.g., rotation, mirror, and defogging) and then concurrently fed into a cascading network for fault diagnosis.
After inputting the parallel images, parallel results would be generated and then a voting decision mechanism was designed for determination of the final result.

In addition to increasing the data diversity, some studies discussed the use of the transfer learning.
Bai et al. \cite{bai2018SPPNet-TL_ins} determined the surface fault of insulators based on Spatial Pyramid Pooling
networks (SPP-Net) with transfer learning.
In the experiment, the model was first trained by the ImageNet dataset which contained among 1.2 million training samples.
Then, the same model was further trained (also denoted as fine-tune) by the small dataset with insulator fault.
Miao et al. \cite{miao2019TS-FT_ins} introduced a two-stage fine-tuning strategy in SSD network to detect insulators.
Two kinds of insulator dataset were prepared in the proposed method: basic dataset and specific dataset.
The former contained aerial images with various types of insulators in different background, which has large quantity.
The later comprised images with the specific insulator in the specific background (e.g., porcelain insulator in forest background), which has little images. 
The implementation of fine-tuning stage 1 was similar to \cite{bai2018SPPNet-TL_ins}.
But instead of ImageNet dataset and small insulator dataset that mentioned in \cite{bai2018SPPNet-TL_ins}, they used COCO dataset and the basic dataset.
In the fine-tuning stage 2, the detection model was further trained by using a specific dataset.
Furthermore, the specific dataset can be replaced according to different engineering applications.
The experiments illustrated the enhanced performance of the proposed strategy compared to the traditional fine-tuning.

Recently, a novel technology called weakly supervised learning was proposed to combat with the data insufficiency that opens a new research issue in the inspection image analysis.
Lee et al. \cite{lee2017weakly_line} segmented power conductors in pixel-level by using data with image-level annotations.
A sliding window combined with CNN was utilized to classify each sub-window of an aerial image into two image-level categories: conductor and background.
If the sub-window was classified as conductor, the bilinear interpolation was applied to up-sampling this sub-window for obtaining the area of conductors.
In the experiment, 4000 images with 128$\times$128 (the size of sub-window) and 200 images with 512$\times$512 were used to train and test the proposed method.
Notice that a real-world can separated into several sub-images as the training samples.
Results with 81.82\% recall rate illustrated the effectiveness of this weakly supervised learning method.

\subsubsection{Improve deep learning method based on domain knowledge of power lines inspection}
The detection and diagnosis tasks of power line components have some contrasts compared to the common task.
(e.g., some faults need to be identified by two stages object detection)
These unique characteristics also can be denoted as domain knowledge in the power lines inspection.
In recent years, few attempts have been made to improve the exiting deep learning method based on this domain knowledge, which makes it more suitable for the data analysis of power lines inspection. 
Jiang et al. \cite{jiang2019EL-MLP} concentrated on the detection procedure of the insulator fault.
The traditional fault diagnosis algorithm was usually a two-stage object detection procedure, which first detected the component and then detect the fault on the component region.
Authors pointed out that the performance of the traditional procedure depending on the effect of component detection, for example, once the component was missing detected, the fault identification can not be achieved.
Therefore, they improved the procedure and proposed a fault diagnosis method based on the ensemble learning with multi-level perception.
They applied SSD to detect the missing-cap in three different input images: original aerial image, multi-insulator image and single-insulator image.
Then, the final result can be filtered by using an improved ensemble learning method.
In the experiment, the improved procedure showed higher accuracy (92.3\%) compared to the traditional procedure (89.1\%) that verified the effectiveness of the proposed method.

Similar to \cite{jiang2019EL-MLP}, Chen et al. \cite{chen2019SOFCN} also discussed the improvement of the fault diagnosis procedure.
A fault detection method of insulators was proposed based on Second-order Fully Convolutional Network (SO-FCN).
They inserted an image filtering operation into the traditional two-stage detection procedure.
The improved procedure consisted of three main steps: the first order FCN was applied to obtain the initial segmentation result of insulator region, morphological reconstruction filtering was performed to remove the false identification, and the second order FCN was employed to detect the missing-cap fault.

Recently, Sampedro et al.\cite{sampedro2019UpNet_CNN_ins} introduced a novel strategy for missing caps detection, which transferred the object detection problem into semantic segmentation problem.
The insulator was conducted by two elements including caps and connectors that were tightly interlocked.
The authors segmented the caps and connectors from an insulator string, and generated a mask image where the pixels belonging to caps were changed to green and the regions of connectors were changed to red.
In this mask image, the detection of missing-cap was transferred to detecting the absent green region.
Moreover, a large number of fault samples can be synthetically produced by randomly removing the green region in the mask image. 
In the experiments, totally 2400 training samples were generated from 160 original images.

In addition to the optimization of fault diagnosis scheme, Xiang et al.  \cite{xiang2018Modified_Fst_ins} improved the deep learning network itself.
They proposed an modified Faster R-CNN to detect the external force damage (e.g., engineering vehicles) of power lines.
According to the characteristics of the engineering vehicles images, for example, the object size, object shape and background, authors modified the Faster R-CNN structure in the feature extraction and classification parts.
In feature extraction, a shallower convolutional neural network was utilized for extracting the high-resolution features .
In feature classification, one convolutional layer was added after the Region of Interest(RoI) pooling layer in order to learn the region-wise features that were suitable for RoI.
These improvements enhanced the ability of the detection network and the advantages of the proposed method (89.93\%) was verified compared to the traditional Faster R-CNN (89.12\%).

\subsubsection{Remarks}
\mbox{Table. \ref{tab:summary_DL}} provides the valuable information of deep-learning-based researches in inspection data analysis, which includes the literature characteristic, inspection item, method, size of total data, pixel size of image, and the core idea of the research.

Although a number of works utilized deep learning methods to analyze inspection data in the past two years, research in this area is still in its early stages.
These works mainly applied existed deep framework (e.g., Faster R-CNN, SSD and YOLO) in a specific inspection item.
More attention should be paid to the improvement of deep learning methods for inspection data analysis instead of direct utilization.
Some primary attempts have been made in this area and several following issues are rose:
For deep feature extracting, which data can be extracted is an important question.
Text with rich information of power lines inspection can be further concerned.
For network cascading, it is worth studying how to solve the coupling problem between object detection stage and fault identification stage, especially the situation of the fault identification can not complete when the object detection fails.
A meticulous designed procedure may be helpful when the object region is miss or wrong detected. 
For data insufficiency, as can be seen in \mbox{Table. \ref{tab:summary_diag}}, a majority of studies used hundreds or thousands samples for experiments that is typically not enough to train a high performance deep learning model.
Few-shot learning is a hot-spot research and some novel methods have been proposed outside the area of power lines inspection.
It is worth trying to apply these state-of-the-art methods to solve the lack of data.
To improve the deep learning method based on domain knowledge in power lines inspection, the characteristic of inspection items need to be further investigated.
Not only the inspection image should be concerned, the information in the whole inspection procedure is also valuable such as the landform, date, weather, and flight record.
Multi-modal learning may be a good choice to handle such complex information.

In addition, we also find that even though the camera can capture high resolution image on UAV (e.g., 4000$\times$3000), the pixel size used for deep learning model training and testing is still small (e.g., 300$\times$300).
How to effectively employ the deep learning method under the situation of large pixel size or limited computation resource is another issue that needs to be further addressed.
Some researches in this area would be helpful to bridge the gap between laboratory work and real-world application.
Besides, the research on deep learning application in power lines inspection will be promoted if some open datasets are provided.

\subsection{A basic conception of inspection data analysis system based on deep learning}

To build an intelligent analysis system of inspection, the following steps should be considered: 
1) process the inspection data for storage and model training using three main approaches including data cleaning, data labeling and data augmentation (Section V.C.1$\sim$V.C.3). 
2) design the component detection method (Section V.C.4). 
3) design the fault identification method (Section V.C.5). 
4) train and optimize the deep learning models in detection stage and identification stage by applying cross-validation, model pruning, and model ensemble (Section V.C.6). 

\begin{figure*}[ht]
	\centering
	\includegraphics[width=0.95\linewidth]{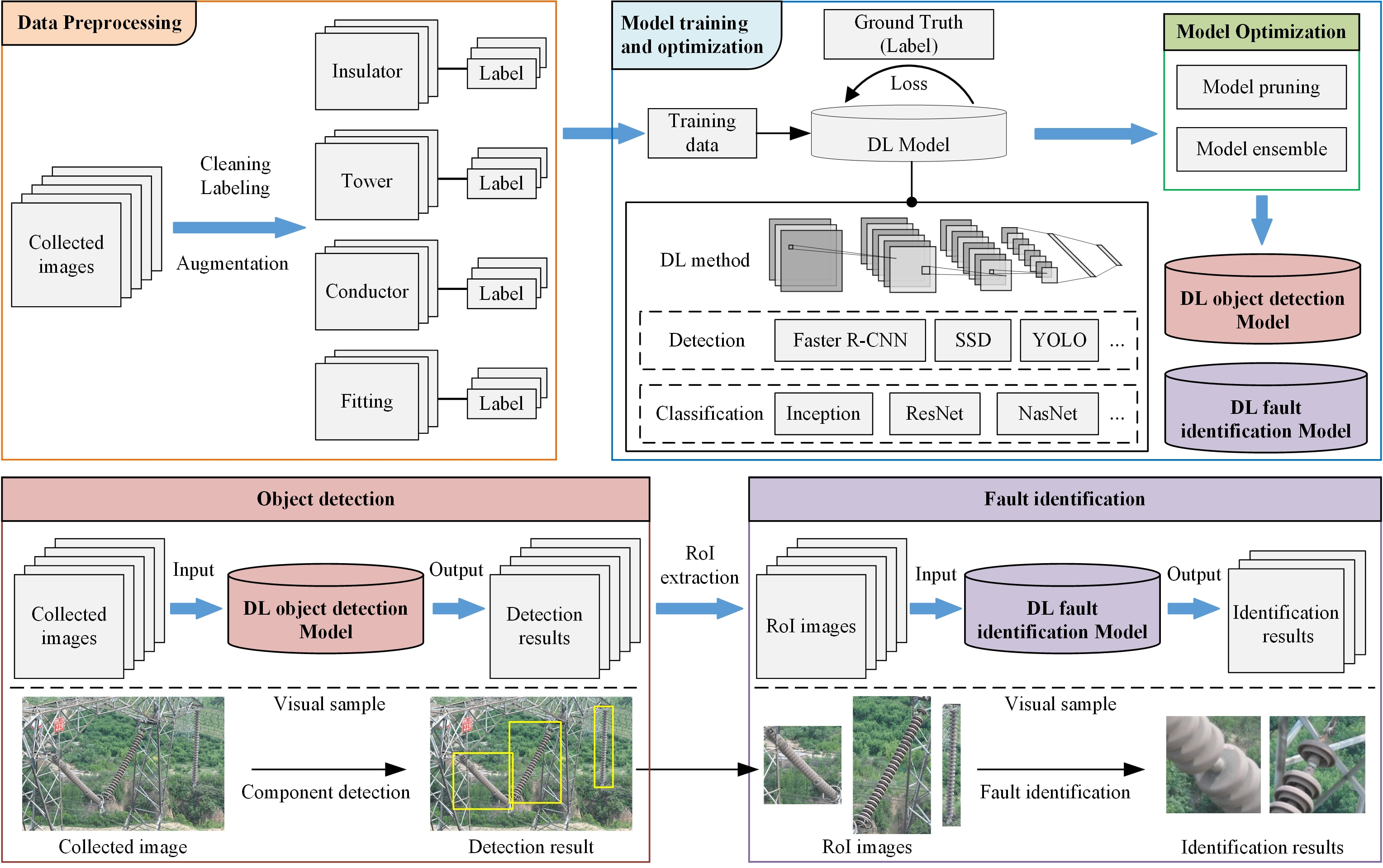}
	\caption{Basic conception of inspection data analysis system based on deep learning}\label{fig:DL_framework}
\end{figure*}

\subsubsection{Data Cleaning}
The aerial images and videos captured from UAVs contain redundant information such as duplicate data, irrelevant data and corrupt data.
These invalid or even harmful data should be filtered out in order to guarantee the model performance and save computation resource.
For achieving this objective, one possible solution is to establish a quality evaluation framework of the inspection data.
Then, the invalid data with low quality can be eliminated.

In order to remove the duplicate data, the similarity comparison method can be applied.
The commonly used approach to compare images is to extract features by descriptors and calculate the squared euclidean distance between these features.
Some hand-craft designed descriptors can be used such as SIFT \cite{Lowe2004SIFT} and DAISY \cite{Tola2008DAISY}.
There are also deep learning based method for similarity comparison, for instance, Siamese Network \cite{Chopra2005Siamese} and 2-Channel Network \cite{Zagoruyko2015Channel2}.
For irrelevant data, we can apply object detection model to detect the power line component.
The aerial image or video without the component is regarded as irrelevant data.
The detection method will be further discussed in subsequent section.
The corrupt data refers to the distorted images caused by UAVs motion, digital compression, and noise interference.
The conventional procedure of filtering the corrupt data is to extract features from aerial image, and then regress these features to a quality score. (The quality here focus on the degree of image distortion which should be distinguished from the quality of inspection data.)
There are some CNN-based approaches that can be applied such as IQA-CNN \cite{2014KangIQA-CNN}, RankIQA \cite{liu2017RankIQA}, BIECON (Blind Image Evaluator based on a Convolutional Neural Network) \cite{kim2016BIECON},  and DIQA (Deep Image Quality Assessor) \cite{kim2018DIQA}.
Once the distorted images are obtained, we can remove or restore according to the application scenario.
For example, not every aerial image in the periodic inspection need to be analyzed, we can remove the distorted images under this condition.
However, in the emergent post-disaster inspection, each image is important and the distorted image should be restored by using CNN-based \cite{sun2015LCNN,nah2017DMsCNN} or GAN-based \cite{kupyn2018DeblurGAN,liu2018XGAN} image restoration approaches.

\subsubsection{Data Labeling}
In order to train and test the deep learning model, the inspection data should be labeled.
The common labeling procedure is to write the image information (e.g., pixel size, object coordinates, and object class) to a file that is independent of the image.
This file also is called annotation file and the file format include TXT, XML, and JSON.
The data labeling can be accomplished manually or semi-automatically.

For manual labeling, two commonly used graphical image annotation tools can be applied: LabelImg \cite{LabelImg} and Labelme \cite{labelme2016}.
In LabelImg, we can click and release left mouse to select a region to annotate the rectangle box which contain the object.
Then, enter the category of the object that exists in the rectangle box.
Finally, the annotation files are saved as XML files in PASCAL VOC format which is commonly used in many dataset (e.g., ImageNet).
The operation of Labelme is similar to LabelImg, but it can achieve more labeling tasks.
Besides the rectangle box, there are many other shapes can be used for image annotation including polygon, circle, line and point.
It is worth noting that the polygon annotation has more detailed contour of the object which can be used for image segmentation task.
The annotation files are saved as JSON file in VOC-format or COCO-format (for COCO dataset).

The procedure of semi-automatic labeling consists of two parts: automatic detection by deep learning models and adjustment by human.
In the first part, a coarse detection model should be trained by using a small part of the entire dataset (manual labeling).
Then, initial annotation files can be obtained by applying the detection model on the rest of the inspection data.
In the second part, the initial annotation file will be adjusted and corrected manually.
There is a semi automatic image annotation tool so-called Anno-Mage \cite{Anno-Mage} can make the this procedure more easier.
A real-time detection model should be prepared and then the image can be detected and adjusted sequentially and interactively.


\subsubsection{Data Augmentation}
Data augmentation is a commonly used technique in deep learning for promoting the performance of the model.
The quantity and diversity of the training data can be augmented by the following approaches: image transformation, image synthesis and GAN-based image generation.

In image transformation, the training sample can be transformed to a new sample by using various image process operations such as rotating, cropping, resizing, shifting, and noising.
These operations can be applied alone or in combination.
Different augmentation strategies will result in different model performance. 
To this end, AutoAument \cite{cubuk2018autoaugment} can be employed to search the optimal strategy.
In addition, there are two implementations of image transformation: before the training and during the training.
The former transforms all the images before the model training that are stored with real-world data together.
The later is more resource efficient that transforms the image in each iteration during model training.

The synthetic image is generated from real-world images by synthesizing the instance image and background image.
The instance image represents the polygon image area of the power line component that the polygon is the contour of the object.
It can be obtained from polygon annotations labeled by hand-craft, or extracted by applying the object segmentation method such as FCN \cite{long2015FCN}, U-Net \cite{ronneberger2015UNet} and Mask R-CNN \cite{he2017mrcnn}.
The background image can be captured from the aerial inspection video of the power line corridor.
By adding the instance image to the background image, a large number of high-quality synthetic image can be obtained.
In addition, some automatic approaches can also generate the synthetic image.
For example, we can applied re-sampling method to synthesize new samples such as Synthetic Minority Over-sampling Technique (SMOTE) \cite{chawla2002smote}, SamplePairing \cite{inoue2018SamplePairing}, and Mixup \cite{zhang2017mixup}.

Recently, the GAN-based image-to-image translation method have opened up possibilities in data augmentation.
The generative model of GAN can generate an new image by inputting an original image.
The image can be transformed to another style such as day-to-night, summer-to-winter, and sunny-to-foggy.
The objects in the image will be also transformed with color, size, and orientation.
With respect to train the generative model, several GAN architectures can be used such as Pix2Pix \cite{isola2017Pix2Pix}, CycleGAN \cite{zhu2017cycleGAN}, and AugGAN \cite{huang2018AugGAN}.

\subsubsection{Component detection}
The component detection in the inspection data analysis refers to obtain the position and the category of the power line component in the aerial image.
In addition, the position information can be represented by the coordinates of rectangular box or polygonal box.
The polygonal box is used in the segmentation task which can also acquire the location of the component in a more meticulous way.
There are two major goals for component detection: 
1) collect key frame from aerial videos that have power line components.
2) crop the component region from the original image for further fault identification.

Given the tremendously rapid evolution of deep learning, there are many successful detection networks that can be applied in the inspection data analysis.
Two main indicators should be considered in selecting these networks for inspection data analysis in different applications: accuracy and speed.
For instance, detection speed is the most important performance indicator in the post-disaster inspection.
But in the long-period inspection task, the electrical company more concentrates on the detection accuracy.
Every detection network aims at detecting the object fast and precise.
However, accuracy and speed are generally contradictory, for example, most high accuracy networks have corresponding high computational cost.
In most case, the two-stage DL-based detection method has higher accuracy than the one-stage method, but in contrast, the former has lower detection speed.
Another factor that affects the performance of the DL-based detection method is the basic network.
Therefore, different combinations of the detection scheme and the basic network have diverse performances.
Recently, a guide was presented by Huang et al. \cite{huang2016modelzoo} for selecting the DL-based detection method that achieves the appropriate performance for a given application.
We can also refer to leaderboards of several large-scale dataset (e.g, COCO and ImageNet) to look for favorable detection methods.
For reference in this paper, two suggested combinations can be applied in the object detection of inspection data.
Concerning applications with high accuracy requirements, the combination of Faster R-CNN with NasNet\cite{pham2018enas} is an exceptional selection.
With respect to the application that requires low computation cost, SSD with ResNet-FPN \cite{lin2017retinanet} can be applied that it can calculate at high speed with the acceptable accuracy.

Furthermore, the DL-based segmentation method (e.g., FCN \cite{long2015FCN}) and DL-based multi-task method (e.g., Mask R-CNN \cite{he2017mrcnn}) can play the role as the detection method does.
But it requires pixel-level annotations which means more labor costs should be paid for.

\subsubsection{Fault identification}
In fault identification, the component region should be cropped first from the original aerial image based on results of component detection stage.
Then, the identification method can be performed on the cropped image.
This two-stage pipeline has following main advantages: 
1) reduce the search range that can improve the accuracy and speed.
2) design component-specific identification methods and perform them on corresponding component region.
Which means there is no need to perform all identification methods in an input image.

The fault identification task in power lines can be summarized into three categories: generic object detection task, generic classification task, and fault-specific task.
In generic object detection task, the fault identification can be regarded as the location of fault regions.
For example, the missing-cap fault of insulators will be determined by detecting the disappeared part of the insulator string.
Similar to missing-cap, many other faults can also be identified by means of object detection such as bird's nest and foreign body.
Therefore, DL-based detection methods mentioned above (e.g., Faster R-CNN with NasNet) can be utilized to accomplish these tasks.

Most identification tasks of surface faults are generic classification tasks due to their irregular fault range and degree.
By using the DCNN (e.g., ResNet), we can classify different surface faults such as flashover of insulators, icing of insulators and corrosion of towers.
Compared to detection tasks that identify the fault by determining the presence or absence of the fault region, the classification task aims at identifying the fault condition level of components.

The remaining faults are difficult to identify by directly detecting or classifying such as broken strand of conductors and vegetation encroachment.
Identification tasks for these faults are summarized as fault-specific tasks.
In order to accomplish these tasks, it is necessary to design special identification methods for different faults.
For instance, a possible solution for broken strand diagnosis is as follow: 
First, segment the contours of conductor in the detected conductor region which is obtained from detection stage.
Then, design hand-craft rules to determined which contour is the broken strand according to the characteristics of the fault.

\subsubsection{Model training and optimization}
After the data are processed and the detection and identification methods are configured, it is essential to obtain available models for real-world applications by model training and optimization. 
In this paper, the DL method refers to the conceptual network architecture while the DL model represents the network that has actual parameters (e.g., weights of the neuron) and can be implemented on a real-world platform.

The model training defined as a procedure of updating parameters with back propagation given the initial model with initial parameters.
There are two frequently-used techniques for model training: fine-tuning and cross-validation.
Fine-tuning is an implementation of transfer learning, where a previously-trained model is utilized as an initial model and then its parameters are adjusted for a new dataset \cite{sze2017survey_DL}. 
Compared with learning from scratch (or training without fine-tuning) that the model with random parameters is used as the initial model, the fine-tuning makes the training process of learning representations more simpler and acquires higher accuracy \cite{yosinski2014transferable}. 

The cross-validation is a widespread technique for combating the over-fitting and making full use of the available data \cite{domingos2012Tips_ML}.
In real-world applications, it is common to split the data into two parts: a part for training and another for validation.
These two subsets of data can be denoted as training set and validation set respectively.
The basic form of cross-validation is k-fold cross-validation, where the data is first split into k equally sets.
Then, subsequently $k$ iterations of training and validation are performed.
In each iteration, a set is held out as validation set and the rest $k-1$ sets are used to train the model.
Finally, we can obtain $k$ well-trained models and corresponding results.
The reliable performance of the method can be acquired by averaging these results and then guides the adjustment of method settings.
In addition, optimal $k$ in k-fold cross-validation is between 5 and 10 that was mentioned by Arlot et al. \cite{arlot2010survey_cross-validation}.

With respect to model optimization, there are two widespread optimized directions: saving the computational cost by model pruning and improving the accuracy by model ensemble.
It is widely-recognized that DL methods are typically over-parameterized in order to train a high performance model with stronger representation power, which leads to high computational cost \cite{sze2017survey_DL}.
As a remedy, the model pruning can remove a set of redundant parameters and its procedure consists of three stages:
1) train an over-parameterized model,
2) prune the redundant parameters of the model according to a certain criterion,
3) fine-tune the pruned model to maintain the original accuracy  \cite{Li2017Pruning,luo2017thinet,liu2018rethinking,huang2018sparse_structure_selection}.

In real-world applications, there are substantial remarkable methods can be employed which yields the difficulty of selection.
To this end, the model ensemble provides a solution of combining multiple methods to obtain better performance \cite{sagi2018ensemble_survey}.
There are two mainstream ensemble methods that have been widely used in classification and object detection tasks: boosting and bagging.
In boosting, the learner (e.g., ResNet) is trained in sequence that each learner depends on the previous learner.
Particularly, each new learner focus on samples the previous ones tended to get wrong.
In bagging, learners are trained independently and parallel, and then the predictions of all learners are combined according to a deterministic average process (e.g., voting) \cite{domingos2012Tips_ML}.
Recently, the model ensemble has been applied to the inspection task that a bagging-like ensemble method was used to detect insulator fault introduced by Jiang et al. \cite{jiang2019EL-MLP}

\section{Challenges and open research issues}
Despite the recent promising results reported in the literature, the adoption of deep learning in image analysis of power lines inspection is still in its infancy and cannot yet satisfactorily address several long-standing challenges.
In this section, we discuss some crucial issues and promising research directions with special attention paid to highlight their challenges and potential opportunities.

\subsection{Data quality problems}
Although the application of UAVs has greatly reduced the workload of inspectors, it has also brought huge amounts of daily data. 
It is an emerging issue to make full use of these inspection data to achieve automatic data analysis.
High-quality data is the guarantee of high performance of analysis methods which are based on machine learning technology.
However, there are four main problems in current inspection data: 
%
\begin{itemize}
	\item \textbf{High labor-cost of data labeling}. 
	Until now, most of the analysis methods are based on supervised learning that are rely heavily on manual annotations.
	But such a large amount of data in power lines inspection requires professionals to spend a lot of time labeling.
	As mentioned by Nguyen et al. \cite{2018DL_inspection_review}, a person needs almost one hour to label 40 images.	
	\item \textbf{Class imbalance}.
	Different components in power lines have different quantities and their faults occur with different frequency.
	For instance, the number of fittings is much larger than the tower and the possibility of failure is also higher.
	In addition, the time accumulation is not enough since the UAV inspection has only been developed for few years. 
	In extreme cases, some categories even have no training data such as tower collapse for specific area.
	These factors lead to class imbalance (also known as long-tailed distribution) that makes the model perform poorly on those categories with insufficient data.
	\item \textbf{Intra-class variations}.
	The problem of intra-class variations is similar to class imbalance in a sense that affects the model performance.
	In real-world application, each category of power line components can have many different object instances, and they possibly have diverse combinations of different characteristics such as color, shape, texture, size, and material.
	Furthermore, the various imaging conditions caused by the changing environment would impact the object appearance even according to the same instance. 
	The changing environment includes the day-to-night changing, weather conditions, photographing orientation and distance, background, occlusion etc.
	In other words, intra-class variations have two manifestations: diverse object instance and complex background.
	\item \textbf{Multiple data sources}.
	In this paper, we only focus on the visible image that is widely used in power lines inspection.
	There are also data from some other sources such as thermal images, ultraviolet images, laser scanner data, and text data which contains flight information.
	How to effectively use these multi-modal data to accomplish the condition identification of the power lines is a challenging problem.
\end{itemize}

These problems in inspection data limit the application of the analysis method for power lines inspection.
In order to offer the potential solutions for the aforementioned challenges, we provide the following research directions.

\subsubsection{\textbf{Weakly supervised object detection}}
Weakly supervised object detection (WSOD) plays a crucial role in relieving human involvement from object-level annotations, and aims at using image-level labels to train an object detector.
If the labeler only needs to care about what the object is in the image without paying attention to its position, it will greatly speed up the labeling process and save a lot of labor costs.
Until now, there is only one work concerning about applying weakly supervised learning into power lines inspection field which attempts to detect conductors by using image-level class labels \cite{lee2017weakly_line}.
There are many other novel WSOD methods \cite{bilen2016WSDDN,tang2018PCL,tang2017OICR,wei2018ts2c} worth trying in components detection and fault identification.
 
\subsubsection{\textbf{Automatic image generation}}
To deal with the problems of class imbalance and intra-class variations, automatic image generation is a very promising approach.
This approach generates rare data by pasting or converting.
In pasting, the demand object should be extract by segmentation network (e.g., Mask R-CNN\cite{he2017maskrcnn}, U-Net\cite{ronneberger2015UNet}, and DeepLab\cite{chen2018deeplabv3}) first, and then paste the object region to the background image.
Few works utilize pasting to generate inspection data including insulator \cite{chang2018Synthetic_ins} and its fault \cite{tao2018ILN_DDN_ins,sampedro2019unet_ins}.
It should be notice that the pasted rule needs refined design in order to obtain realistic data.
In converting, the new image is generally converted from the old one by using the Generative Adversarial Networks (GANs) \cite{goodfellow2014GAN}.
An example is shown in work \cite{lu2019trasfer_ins} which realizes the mutual conversion of normal image and fault image by means of CycleGAN \cite{zhu2017cycleGAN}.
There are some other powerful GANs can be used for image generation such as Pix2Pix \cite{isola2017Pix2Pix} and AugGAN \cite{huang2018AugGAN}.
In this direction, how to bridge the reality gap is important for generating the high quality and realistic synthetic data\cite{cong2019_Image_Harmonization, tremblay2018DR}.

\subsubsection{\textbf{Multi-modal object detection}}
To take advantage of multiple data sources, the technology of multi-modal object detection can be applied.
The objective is to fuse the information from different modalities to achieve a more discriminant detection method.
In the exiting works of power lines inspection, few researchers attempt to make use of multiple data sources.
Zhao et al. \cite{zhao2019frcnn_ins} discussed that it is feasible to detect insulators in visible images by using the model trained from infrared images.
Maeda et al. \cite{maeda2018CMDELM-LRF} extracted deep features from visible images and text to identify the deterioration level of tower.
Jalil et al. \cite{jalil2019multimodal} applied multi-modal imaging which integrated the infrared and visible images for fault identification of the power line component.
Nevertheless, the components or faults detection based on multi-modal data is still in its early stage.
In this direction, the questions of "what sources to fuse" and "how to fuse" are important for designing the multi-modal based method. 
Some works in generic tasks such as pedestrian detection \cite{guan2019_pedestrian_det}, car detection \cite{chen2017_car_det} and medical image analysis \cite{xu2019_medical}, can be used as as references.

\subsection{Small object detection}
\begin{figure*}[ht]
	\centering
	\includegraphics[width=0.95\linewidth]{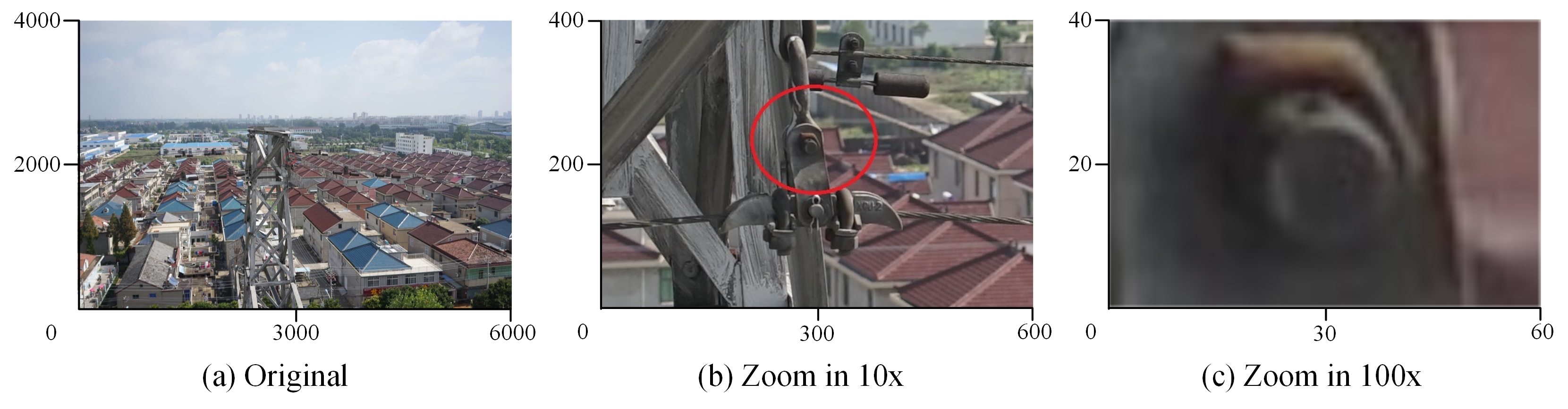}
	\caption{An example of small object. The fault in the image is missing pin of fittings.}\label{fig:challenge_small_obj}
\end{figure*}

There are many small components in the inspection image such as the fitting and conductor.
\mbox{Fig. \ref{fig:challenge_small_obj}} illustrates an example of the small object.
The fault in this sample is missing pin of fittings.
As can be seen in the image, the pixel resolution of the component region is merely 60$\times$40 in the whole image with 6000$\times$4000.
The object is already very small, however, the high resolution image should be resized to a smaller resolution (e.g., 300$\times$300) during training that makes many features disappear.
Unfortunately, the pooling and down-sampling operation in the deep network makes this problem worse.

To detect small objects in aerial inspection images, there are three potential solutions:
The first is to directly enlarge images to different scales.
An example is provided by Bai et al. \cite{bai2019frcnn_multi} for insulators and dampers detection.
The second is multi-stage detection strategy which utilizes the contextual information.
The component with large size is firstly detected and cropped as ROIs, and then small objects are located in these ROIs.
This solution also has been applied for the identification of insulator fault \cite{jiang2019EL-MLP,ling2018Fst-Unet,chen2019SOFCN}.
The third is to improve the deep neural network by fusing the features in different scales.
Han et al. \cite{Han2019MYOLO_AM_ins} add three branches into YOLOv3 network for insulator detection.
Besides that, there are some other fusion methods in generic can be applied such as Feature pyramid networks (FPN) \cite{lin2017fpn}, Top-Down Modulation (TDM) \cite{shrivastava2016tdm}, and Reverse connection with objectness prior networks (RON) \cite{kong2017ron}.
The key in this solution is how to make use of the rich information of small objects in low-level feature maps of shallow convolutional layers.

\subsection{Embedded application}
Due to the increasing demands of high performance computation, reducing data transmission, and achieving highly efficient inspection, it is necessary to accomplish some processes of the analytic system on site (also means on-board the UAVs).
Even though some of the current embedded computing devices, such as NVIDIA Jetson TX2, can undertake complex image processing tasks including light DCNN, they still can not handle the high performance analysis methods.
Therefore, how to make inspection data analysis more efficient with short computing time and small memory usage is an important issue for practical engineering.
In this research direction, the technologies of model compression and acceleration \cite{cheng2017comprossion_survey} can be applied that include model pruning \cite{liu2017model_pruning}, network quantization \cite{hubara2017network_quantized}, network decomposition \cite{liu2015network_decomposition}, knowledge distillation \cite{hinton2015knowledge_distillation}, and lightweight network design \cite{howard2019mobilenetv3}.


\subsection{Evaluation baseline}
The evaluation baseline refers to the evaluation metrics and the standard dataset, which can offer a public platform for researchers and facilitate related practice.
Currently, the evaluation metrics used in researches of inspection image analysis are diverse, for instance accuracy rate, precision, true positive rate, false alarm rate, detection rate etc.
Even in the same evaluation metric, the definition may be different especially the accuracy rate.
In addition, the available public datasets of power lines inspection are not enough to build a comprehensive standard dataset that can well evaluate the performance of an analysis system.
As for building an evaluated baseline, the generic and successful dataset such as ImageNet \cite{russakovsky2015ILSVRC} and COCO \cite{lin2014cocodataset}, can provide some experience.
When constructing the dataset, many factors should be considered, for instance the component category, fault type, labeled rule, flight environments, and size of samples. 
We deem that an successful evaluation baseline can facilitate the studies and applications of power lines inspection.

\section{Conclusion}
In this paper, we have provided a comprehensive review of inspection data analysis in power lines. 
The latest developments have been summarized and the key characteristics of these researches have been discussed. 
Firstly, studies on power line component detection in inspection images are reviewed from the perspective of insulator, tower, conductor, and fitting.
Then, the literature survey of power line fault identification is conducted in a fault-specific way including surface fault of insulator, missing-cap of insulator,  tower corrosion, bird's nest, broken strand, foreign body, vegetation encroachment, broken fitting, and missing-pin of fitting.
Next, a thorough review about deep learning related works in the area of data analysis of power lines inspection is introduced.
These articles are categorized into five groups including direct utilization of existing frameworks, deep feature extraction, network cascading, data insufficiency issue, and improvement based on domain knowledge.
Further, a basic conception of inspection data analysis system which is mainly based on deep learning technology is proposed.
This system consists four parts: data preprocessing, component detection, fault diagnosis, and model training and optimization.
Finally, we discuss the challenges and propose future research directions from the prospective of data quality, small object detection, embedded application, and evaluation baseline.
Inspection data analysis in power lines is still an emerging and promising research area.
We hope that this review can provide a complete picture and deep insights into this area for researchers who are interested in developing a automatic analysis system of power line inspection data using deep learning technology.

\ifCLASSOPTIONcaptionsoff
  \newpage
\fi

\bibliographystyle{IEEEtran}
\bibliography{insulator}

\begin{IEEEbiography}[{\includegraphics[width=1in,height=1.25in,clip,keepaspectratio]{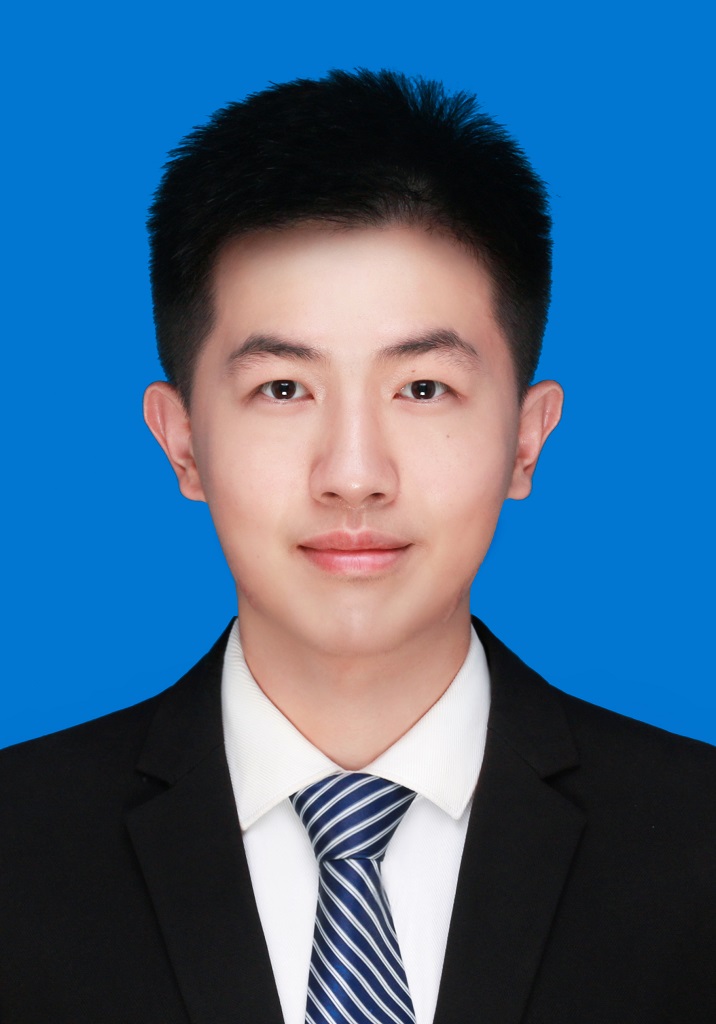}}]{Xinyu Liu} received the B.S. and M.S. degree at Fuzhou University, Fujian, China, in 2016 and 2019 respectively. He is currently pursuing the Ph.D. degree in power system and its automation in the Fuzhou University. His research interests include image processing, deep learning and condition monitoring of power lines.
\end{IEEEbiography}
\begin{IEEEbiography}[{\includegraphics[width=1in,height=1.25in,clip,keepaspectratio]{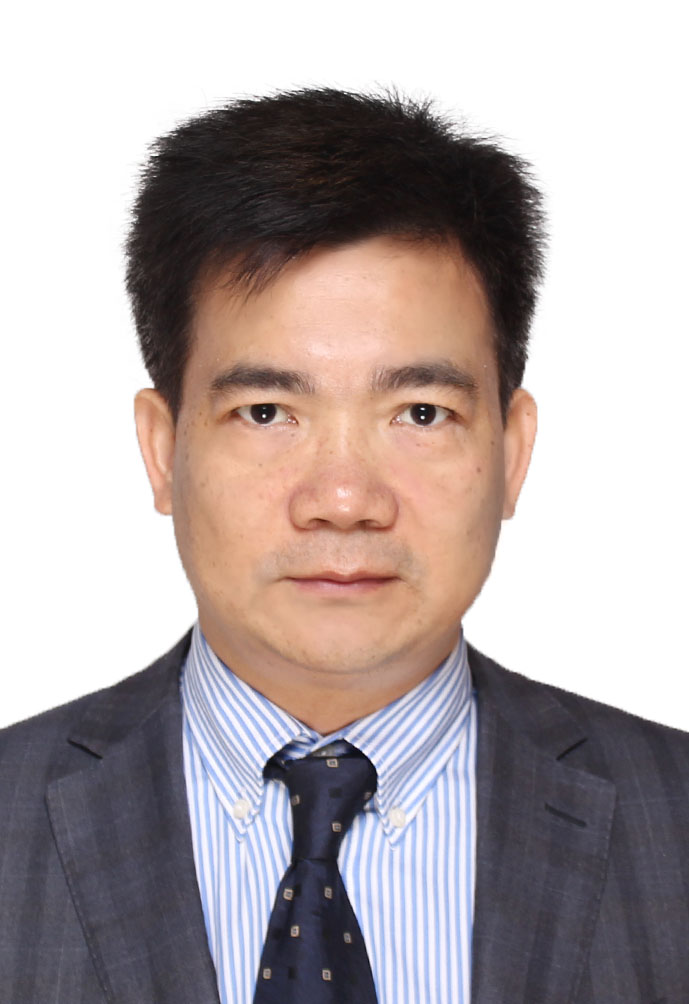}}]{Xiren Miao} received the B.S. degree at Beihang University, Beijing, China, in 1986, and received the M.S. and Ph.D. degrees from the Fuzhou University, Fuzhou, China, in 1989 and 2000. He is currently a Professor with the College of Electrical Engineering and Automation, Fuzhou University. His research interests include electrical and its system intelligent technology, on-line monitoring and diagnosis of electrical equipment.
\end{IEEEbiography}
\begin{IEEEbiography}[{\includegraphics[width=1in,height=1.25in,clip,keepaspectratio]{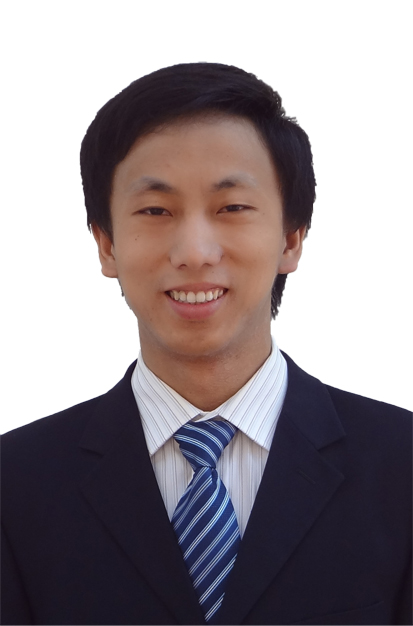}}]{Hao Jiang} received the B.S. and Ph.D. degrees at Xiamen University, Fujian, China, in 2008 and 2013. He is currently a Associate Professor with the College of Electrical Engineering and Automation, Fuzhou University. His research interests include artificial intelligence and machine learning. 
\end{IEEEbiography}
\begin{IEEEbiography}[{\includegraphics[width=1in,height=1.25in,clip,keepaspectratio]{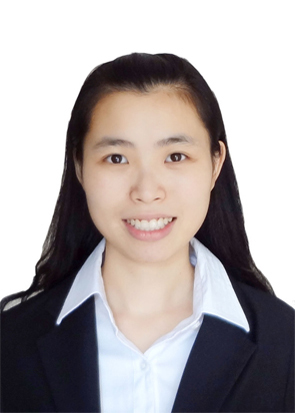}}]{Jing Chen} received the B.S., M.S. and Ph.D. degrees from the Xiamen University, Fujian, China, in 2010, 2013, and 2016 respectively. She is currently a lecturer with the College of Electrical Engineering and Automation, Fuzhou University. Her research interests focus on intelligent fault diagnosis and artificial intelligence.
\end{IEEEbiography}

\end{document}